\documentclass[11pt,a4paper]{article}

\usepackage[utf8]{inputenc}
\usepackage[T1]{fontenc}
\IfFileExists{lmodern.sty}{\usepackage{lmodern}}{}
\usepackage[margin=1in]{geometry}
\usepackage{parskip}

\usepackage{amsmath,amssymb,amsfonts,mathtools,bm,amsthm}

\usepackage{booktabs,array}
\usepackage{enumitem}
\usepackage{xcolor}
\usepackage{caption}
\usepackage{float}

\usepackage{graphicx}
\usepackage{tikz}
\usepackage{algorithm}
\usepackage{algpseudocode}
\usepackage[authoryear,round]{natbib}
\usepackage[hidelinks]{hyperref}
\IfFileExists{lmodern.sty}{\usepackage{microtype}}{\usepackage[expansion=false]{microtype}}

\usetikzlibrary{arrows.meta,positioning,fit,calc,shapes.geometric,backgrounds}

\newcommand{\incfig}[2][0.8\textwidth]{%
  \IfFileExists{#2.pdf}{\includegraphics[width=#1]{#2.pdf}}%
  {\IfFileExists{#2.png}{\includegraphics[width=#1]{#2.png}}%
  {\IfFileExists{#2.jpg}{\includegraphics[width=#1]{#2.jpg}}%
  {\IfFileExists{#2}{\includegraphics[width=#1]{#2}}%
  {\fbox{\parbox[c][0.22\textheight][c]{#1}{\centering\ttfamily\small
     Placeholder for figure\\[2pt]\detokenize{#2}\\[4pt]%
     \normalfont\itshape (image file not found at compile time)}}}}}}%
}

\newcommand{\vect}[1]{\mathbf{#1}}          
\newcommand{\mat}[1]{\mathbf{#1}}            
\newcommand{\ten}[1]{\underline{\bm{#1}}}   

\newcommand{\R}{\mathbb{R}}

\DeclareMathOperator{\Var}{Var}

\newcommand{\softmax}{\operatorname{softmax}}
\newcommand{\loss}{\mathcal{L}}              
\DeclareMathOperator{\rank}{rank}
\DeclareMathOperator{\vecop}{vec}
\DeclareMathOperator{\fold}{fold}
\DeclareMathOperator{\unfold}{unfold}
\DeclareMathOperator{\Contract}{Contract}

\DeclareMathOperator{\id}{id}
\DeclareMathOperator{\concat}{concat}

\newtheorem{remark}{Remark}
\newtheorem{definition}{Definition}

\title{Automatically Differentiable Nonlinear Tensor Networks (ADNTNs) 
for Exponential Parameter Compression of Deep Neural Networks}
\author{Andrzej Cichocki and Micha{\l} Wi{\k{e}}tczak \\[2pt]
  \small Systems Research Institute of Polish Academy of Science 
  and \\ \small Warsaw University of Technology, Poland \\
  \small \texttt{Andrzej.Cichocki@pw.edu.pl}}

\begin{document}
\maketitle


\begin{abstract}

The relentless growth of large-scale deep neural networks (DNNs) and large
language models (LLMs)---now routinely exceeding hundreds of millions of
parameters---creates a fundamental tension between model expressivity and
practical deployability.  Although overparameterisation facilitates optimisation
and promotes generalisation, it imposes severe costs at inference time: immense
memory footprints, high memory-bandwidth pressure, costly hardware accelerators,
and substantial energy consumption per query.  A principled solution to this
challenge requires methods that compress parameters \emph{structurally}, by
capturing the intrinsic low-dimensional geometry of trained weight tensors,
rather than by heuristic pruning or uniform quantisation.

We introduce \emph{Automatically Differentiable Nonlinear Tensor Networks}
(ADNTNs), a family of compact, hierarchical weight generators that compress the
parameters of dense, fully connected, convolutional, and attention layers of deep neural networks by several orders of magnitude without dramatic reduction of accuracy.
Each ADNTN replaces a large weight tensor by a small, trainable tensor-network
program whose core tensors are optimised end-to-end through reverse-mode
automatic differentiation (AD).  ADNTNs generalise both low-rank adaptation
(LoRA) and classical tensor factorisation: instead of a single rank-$r$ update,
the weight tensor is built through a structured hierarchy of small cores,
\emph{learnable nonlinear activations}, and optional lateral disentangling
core tensors. 

 We propose to employ three complementary, multilayered--hierarchical decoders---Tree Tensor Networks (TTNs),
augmented TTNs (aTTNs) with boundary disentanglers, and the Multi-scale
Entanglement Renormalisation Ansatz (MERA)---and show how their nonlinear
extensions overcome the multilinear-rank limitations of classical decompositions, especially, 
such as tensor train (TT/MPS), Tucker, and Canonical Polyadic Decomposition (CPD).

  Inserting learnable activations between
contraction layers of tensor networks breaks the fixed cut-rank bounds of multilinear networks,
letting the generated weights occupy far richer low-dimensional manifolds than
any purely linear factorisation can reach.  Methodologically, we contribute a
unified forward--adjoint calculus for the TTN, aTTN, and MERA generators---with
explicit reverse-mode equations for pre-activation adjoints, parent-state
adjoints, and contracted-environment core gradients.

Proof-of-concept experiments on compression of selected large layers of AlexNet and VGG-16 for  CIFAR10 datasets demonstrate per-layer parameter-compression ratios ranging from roughly  $2{,}000\times$ to  
$430{,}000\times$ in the studied settings.  These results suggest that ADNTNs
are a promising and mathematically interpretable route toward much smaller neural
networks, provided that topology, optimisation, contraction schedules, and
hardware kernels are designed together.
 Because the proposed generators are differentiable
program, the framework supports arbitrary task-aware objectives and integrates naturally with modern AD libraries such as PyTorch, JAX,
and TensorFlow.
\end{abstract}


\section{Introduction}
\label{sec:intro}

Modern state-of-the-art deep neural networks (DNNs) and large language models
(LLMs) routinely contain hundreds of millions to hundreds of billions of
parameters.  They are trained on large distributed hardware clusters and then
deployed in latency-sensitive applications spanning image classification,
medical screening, interactive dialogue, document retrieval, and mobile
assistance.  This scale promotes expressivity and facilitates optimisation, but
it creates severe practical obstacles at inference time: a large model demands
high memory capacity and bandwidth, expensive hardware accelerators, and
non-negligible energy per query.  As model size continues to grow, the gap
between training infrastructure and deployment constraints widens, making
principled compression an increasingly critical challenge.

A substantial empirical literature shows that many trained networks are
functionally redundant: parameters can often be pruned, quantised, factorised,
shared, or distilled with limited loss of accuracy in the moderate-compression
regime \citep{Blalock2020,Han2016,Ganesh2020}.  Tensor decompositions provide a
structured, mathematically grounded way to exploit this redundancy because they
replace a dense tensor by a network of smaller factors whose parameter count
scales linearly (or near-linearly) in the number of tensor modes rather than
exponentially \citep{KoldaBader2009,Cichocki2016,Oseledets2011,Wang2023}.
The same machinery has long been applied directly to neural-network layers, both
as a post-hoc factorisation of pre-trained weights and as a tensor-network
parameterisation learned end-to-end from data
\citep{Novikov2015,Stoudenmire2016,Orus2014}.

Recent seminal work on automatically differentiable tensor networks (ADTNs) has
demonstrated that brick-wall tensor-network architectures can compress large
fully connected and convolutional layers at remarkable ratios when trained via
automatic differentiation \citep{Qing2025}.  In that scheme, the parameters of a
layer are encoded into a flat, brick-wall network of four-index cores
$\ten{A}^{[k]}\in\R^{d\times d\times d\times d}$, ($d=2$) contracted against a fixed
boundary vector and interleaved with \emph{fixed} ReLU activations; the cores
are then optimised by AD, first against a Frobenius pre-training loss and then
against the task loss (cross entropy).  This is a promising and very influential idea, and it is the
direct starting point for the present work.  However, the brick-wall topology
has two structural limitations that we address here.  First, a brick-wall network
propagates information only \emph{locally}, column by column: to couple two
distant tensor indices it must stack enough layers for their light cones to
overlap.  Capturing the global, long-range correlations of a strongly correlated
layer therefore forces the network to be made very deep---which increases the
parameter count, erodes the compression objective, and reintroduces the
vanishing/exploding-gradient pathologies that motivate compression in the first
place.  A \emph{shallow} brick-wall network cannot compress a highly correlated
layer without discarding critical long-range structure.  Second, the fixed ReLU
nonlinearity has zero derivative on its entire negative branch, which can freeze
information paths through the narrow $\chi=2$ bonds that make the scheme so
compact, leaving large parts of the generator effectively untrained.

\textbf{This paper addresses these limitations directly.}  We propose
\emph{Automatically Differentiable Nonlinear Tensor Networks} (ADNTNs), a family
of hierarchical, tree-structured weight generators that capture multi-scale and
long-range correlations through a logarithmic-depth topology, and that overcome
the multilinear rank restrictions of classical tensor networks by inserting
\emph{learnable} nonlinear activations between contraction layers.  Two design
changes distinguish ADNTNs from the brick-wall ADTN of \citet{Qing2025}.
(i)~\emph{Hierarchical instead of flat connectivity.}  The TTN/aTTN/MERA topology
achieves global information flow in $O(\log Q)$ layers rather than the $O(Q)$
depth a brick-wall network needs to couple distant modes, so a single shallow
hierarchy reaches every index without the depth blow-up that afflicts brick-wall
designs.  (ii)~\emph{Adaptive instead of fixed nonlinearity.}  Replacing fixed
ReLU by learnable self-gated activations (Swish/SiLU with trainable gate,
PELU, and related families) keeps gradients alive across narrow bonds and lets
different parts of the generator interpolate between near-linear smoothing and
hard thresholding.  Together with boundary disentanglers (aTTN) and full
multi-scale disentangling (MERA)---mechanisms that have no analogue in a strict
brick wall---these changes allow the generated weights to lie on manifolds far
richer than those accessible to any purely multilinear or flat factorisation.
As we show in Section~\ref{sec:experiments}, under a matched protocol all three
hierarchical generators outperform a faithfully reproduced brick-wall ADTN
baseline especially in respect to compression rate  with higher or similar accuracy.

\subsection{Motivation and Compression Objectives}
\label{sec:motivation}

The practical case for model compression is straightforward: a model that
achieves the same task performance with far fewer parameters should be potentially 
cheaper to store,
faster to execute, and consumes less energy---properties that are critical for
deployment on edge devices, in data centres operating under energy constraints,
and in latency-sensitive interactive applications.  

Yet compression is a
genuinely multi-objective design problem that seeks to balance several competing
goals simultaneously.

\begin{itemize}[leftmargin=*]
  \item \textbf{Memory reduction:} reduce the number of stored parameters and
        the bytes per parameter.
  \item \textbf{Latency reduction:} reduce wall-clock inference time, not merely
        the formal parameter count.
  \item \textbf{Energy and cost reduction:} reduce accelerator memory traffic,
        arithmetic intensity, and operational cost.
  \item \textbf{Accuracy preservation:} maintain the predictive performance of
        the original model on the target task.
  \item \textbf{Behavioural fidelity:} preserve the teacher (dense)  model's output
        distribution, calibration, intermediate representations, and
        safety-relevant behaviour.
  \item \textbf{Deployability:} produce structures that can be executed
        efficiently by available libraries and hardware.
\end{itemize}

Although all of these objectives are desirable, they can be mutually
contradictory.
Unstructured sparsity may reduce the parameter count but often fail to accelerate
inference on dense hardware.
Very aggressive quantisation may reduce memory bandwidth but damage rare
capabilities.
Domain-specific compression may improve a specialised benchmark while weakening
general-purpose behaviour.

Compression is therefore best understood as a Pareto frontier,
\[
  \bigl(\text{size},\ \text{latency},\ \text{energy},\ \text{accuracy},\
  \text{fidelity}\bigr),
\]
rather than as a single universal ratio.

\subsection{Classical Compression Families}
\label{sec:families}

The principal classical compression families are the following.
\begin{itemize}[leftmargin=*]
  \item \textbf{Pruning:} removing individual weights, channels, heads,
        neurons, filters, or blocks according to a saliency criterion
        \citep{Blalock2020}.
  \item \textbf{Quantisation:} replacing 32-bit or 16-bit floating-point
        parameters and activations by lower-precision formats, commonly 8-bit,
        4-bit, or mixed precision \citep{Dettmers2023,Frantar2023}.
  \item \textbf{Knowledge distillation:} training a compact student to imitate
        the logits, probabilities, hidden states, attention maps, or behaviour of
        a larger (dense) pre-trained teacher \citep{Hinton2015,Ganesh2020}.
  \item \textbf{Parameter-efficient adaptation:} freezing a pre-trained model
        and learning a small structured update, as in LoRA, where
        $\Delta\mat{W} = \mat{B}\mat{A}$ is low-rank \citep{Hu2022}.
  \item \textbf{Low-rank and tensor-network factorisation:} approximating a
        weight matrix by $\mat{W} \approx \mat{A}\mat{B}$ with a small inner
        dimension, or approximating a higher-order weight tensor by a
        Tensor Train (TT/MPS), Tensor Ring (TR), Matrix Product Operator (MPO),
        Tucker, Canonical Polyadic Decomposition (CPD), Tree Tensor Network
        (TTN), or MERA factorisation
        \citep{Denton2014,Lebedev2015,Kim2016,Phan2020,Gusak2019,Qing2025,Gu2025,Javanmard2026,Wang2023}.
\end{itemize}

LoRA illustrates the conceptual transition from dense retraining to structured
low-dimensional adaptation, and has achieved remarkable practical success in
fine-tuning LLMs.  Tensor-network compression generalises this idea further:
instead of a single rank-$r$ matrix update, a large weight matrix or
convolutional kernel is represented as the \emph{contraction of many small core
tensors}, enabling compression ratios that grow exponentially with the tensor
order.  The principal examples are the TT/MPS, Tensor Ring (TR), Matrix Product
Operator (MPO), CPD, Tucker, TTN, and MERA factorisation families
\citep{Cichocki2016,Cichocki2017,Oseledets2011,KoldaBader2009,Zhao2016,Vidal2007}.
ADNTNs extend this family by making the generator nonlinear and end-to-end
differentiable, unlocking compression regimes that are inaccessible to
multilinear methods.

\subsection{From ADTN to ADNTN: Key Definitions}
\label{sec:adtn_def}

\begin{definition}[ADTN]
An \emph{Automatically Differentiable Tensor Network} (ADTN) is a tensor
network implemented as a differentiable computation graph so that its core
tensors can be optimised by automatic differentiation (AD), with optional
ReLU nonlinearities \citep{Liao2019,LiaoLiu2021,Qing2025}.
\end{definition}

\begin{definition}[ADNTN]
An \emph{Automatically Differentiable Nonlinear Tensor Network} (ADNTN) is an
ADTN in which learnable nonlinear activation functions, normalisation layers,
gating mechanisms, or other differentiable nonlinearities are inserted between
tensor contractions, so that the overall generation map is nonlinear in the
trainable core tensors.
\end{definition}

In a classical tensor network, all contractions are multilinear in the core
tensors, and the generated weight tensor lies in a product of linear subspaces
controlled by the bond dimensions.  In an ADNTN, the contractions remain
differentiable---and hence amenable to end-to-end training by AD---but are no
longer multilinear, because activations such as SWISH or PELU are interleaved with
the contraction layers \citep{Hendrycks2016,Clevert2015}.  This seemingly small
modification has a profound consequence: the algebraic rank bounds that limit
purely multilinear tensor networks (Eq.~\eqref{eq:cut-rank-bound}) no longer
apply, and the generator can produce weight tensors that lie on richer,
curved manifolds.  This modification is especially important when small-radix
($d=2$) cores are used, as discussed in detail in Section~\ref{sec:nonlinearity}.

Within this taxonomy, the brick-wall ADTN of \citet{Qing2025} is the flat,
fixed-ReLU special case of the broader ADNTN family.  ADNTNs generalise it along
two orthogonal axes: the \emph{connectivity} of the generation graph (from a flat
brick wall to a logarithmic-depth hierarchy with optional disentanglers) and the
\emph{nonlinearity} (from a fixed ReLU to learnable, self-gated activations).
The remainder of the paper develops both axes and quantifies their effect.

\subsection{Main Contributions}
\label{sec:contributions}

The main contributions of this paper are as follows.
\begin{enumerate}[leftmargin=*]
  \item \textbf{ADNTN framework.}  We formulate ADNTNs as hierarchical
        nonlinear tensor-network weight generators for dense, convolutional, and
        attention layers, and show how learnable activations between contraction
        layers break the multilinear rank bottleneck of classical tensor
        networks, allowing generated weights to occupy richer low-dimensional
        manifolds.  We position the brick-wall ADTN of \citet{Qing2025} as the
        flat, fixed-ReLU special case of this family and generalise it along two
        axes: hierarchical $O(\log Q)$ connectivity and learnable nonlinearity.
  \item \textbf{Unified forward--adjoint calculus.}  We derive explicit
        reverse-mode AD equations---forward contractions, pre-activation
        adjoints, and contracted-environment gradients---for all three generator
        architectures: TTN, aTTN, and MERA.  These equations make precise what
        any standard AD engine computes when \texttt{loss.backward()} is called
        on an ADNTN, and serve as a reference for custom kernel implementations.
  \item \textbf{Conceptual clarity.}  We prove why a nonlinear generation graph
        is beneficial even when the generated layer is used linearly at inference
        time, and we carefully distinguish between \emph{differentiating} a
        contraction program and making contraction \emph{free}---a conflation
        that has led to overclaims in prior work.
  \item \textbf{Unified notation and initialisation policy.}  We synchronise
        notation for generated tensors, target tensors, adjoints, bond
        dimensions, physical radix, and layer indices across all three
        architectures, and introduce a global initialisation convention
        (He/Kaiming for hidden cores, Xavier/Glorot for output cores,
        near-identity for disentanglers) that is essential for numerically stable
        training of deep hierarchical generators.  This replaces the per-layer,
        two-stage Frobenius pre-training of brick-wall ADTN with a single
        end-to-end recipe shared by all topologies.
  \item \textbf{Proof-of-concept experiments and head-to-head comparison.}  We
        report compression experiments on AlexNet and VGG-16 layers (CIFAR-10),
        documenting per-layer parameter counts, compression ratios from roughly
        $2{,}000\times$ to $430{,}000\times$, and validation-accuracy changes
        relative to the dense baseline, including several cases in which the
        compressed model \emph{exceeds} baseline accuracy.  Under a matched
        protocol, all three hierarchical generators outperform a faithfully
        reproduced brick-wall ADTN baseline.
\end{enumerate}

\subsection{Paper Organisation}
\label{sec:organisation}

Section~\ref{sec:Tensorization} introduces tensorisations of DNN layers.
Section~\ref{sec:architectures} describes the hierarchical nonlinear tensor
network decoders used in the paper.  Section~\ref{sec:nonlinearity} explains
why a nonlinear generator can be beneficial even when the generated DNN layer is
linear or multilinear in the data.  Section~\ref{sec:loss} discusses
reconstruction, task-aware, distillation, regularisation, and quantisation-aware
losses.  Section~\ref{sec:ad} reviews automatic differentiation and
differentiable programming.  Section~\ref{sec:training} derives the AD learning
algorithms.  Section~\ref{sec:experiments} reports implementation details and
proof-of-concept simulations.  Section~\ref{sec:conclusions} concludes and
outlines future directions.

\subsection*{Notation Guide}
\label{sec:notation_guide}

The following typographic conventions are used consistently throughout the
paper.
\begin{itemize}[leftmargin=*]
  \item \textbf{Scalars} are plain italic letters, for example
        $d$, $Q$, $L$, $N_\ell$, $\chi$, $\eta$, $\lambda$, and
        $\varepsilon_{\rm opt}$.
  \item \textbf{Vectors} are bold lowercase, for example
        $\vect{x}$, $\vect{y}$, $\vect{h}$, and $\vect{w}$.
  \item \textbf{Matrices} are bold uppercase, for example
        $\mat{W}$, $\mat{A}$, $\mat{B}$, and $\mat{S}$.
  \item \textbf{Tensors} of order at least three are underlined bold uppercase,
        for example $\ten{W}$, $\ten{K}$, $\ten{V}$, $\ten{U}$, and
        $\ten{C}$.
  \item \textbf{Generated quantities} carry a hat, for example
        $\widehat{\ten{W}}_\Theta$ or $\widehat{\mat{W}}_\Theta$; target
        quantities carry the subscript ``tar'', for example $\ten{W}_{\rm tar}$.
  \item \textbf{Adjoints} are denoted by an overline:
        $\overline{\ten{X}}:=\partial\loss/\partial\ten{X}$ and
        $\bar v:=\partial\loss/\partial v$.  This convention is used for both
        manual formulae and AD-computed gradients.
\end{itemize}

A summary of the principal symbols is given in Table~\ref{tab:notation}.

\begin{table}[H]
\centering
\small
\caption{\textbf{Notation used throughout the paper.}  The table synchronises the
symbols used in the tensorisation, forward-contraction, adjoint, optimisation,
and experimental sections.  In code, indices may be zero-based; in the
mathematical formulae we use $[d]=\{1,\ldots,d\}$.}
\label{tab:notation}
\begin{tabular}{lp{2.7cm}p{8.0cm}}
\toprule
\textbf{Symbol} & \textbf{Type} & \textbf{Meaning} \\
\midrule
$\vect{x}$, $\vect{y}$, $\vect{h}$, $\vect{w}$ & vector & input, label/target, hidden activation, and vectorised weight \\
$\mat{W}$, $\widehat{\mat{W}}_\Theta$ & matrix & dense weight matrix and generated/compressed weight matrix \\
$\ten{K}$, $\widehat{\ten{K}}_\Theta$ & tensor & convolutional kernel and generated/compressed convolutional kernel \\
$\ten{W}_{\rm tar}$ & tensor & tensorised target obtained by padding and folding a dense layer \\
$\widehat{\ten{W}}_\Theta$ & tensor & tensor generated by the ADNTN decoder before reshaping into a layer \\
$\ten{W}^{(\ell)}$, $\ten{Z}^{(\ell)}$ & tensor & state tensor and pre-activation tensor at generator layer $\ell$ \\
$\ten{C}$, $\Theta$ & tensor/set & a generic trainable core and the full set of trainable latent tensors and cores \\
$\ten{V}^{(\ell,k)}$ & tensor & binary expansion core at layer $\ell$, branch/node $k$ \\
$\ten{U}^{(\ell,k)}$ & tensor & two-site disentangler or lateral mixing core at layer $\ell$, position $k$ \\
$\overline{\ten{X}}$ & tensor & adjoint $\partial\loss/\partial\ten{X}$ with the same shape as $\ten{X}$ \\
$\loss$ & scalar & scalar objective optimised by AD \\
$d$, $\chi$ & scalar & physical radix and virtual/bond dimension \\
$Q$, $Q_{\rm bot}$, $N_\ell$, $L$ & scalar & target tensor order, latent bottleneck order, number of modes after layer $\ell$, and generator depth \\
$\sigma$, $\phi_\ell$ & function & base nonlinearity and layer-dependent activation, with $\phi_L=\id$ at the output \\
$\eta$, $\lambda$, $\varepsilon_{\rm opt}$ & scalar & learning rate, AdamW weight decay, and numerical stabiliser in AdamW \\
$D_{\rm in}$, $D_{\rm out}$ & scalar & input and output dimensions of the original layer \\
$P$, $\bar P$, $\rho_{\rm par}$ & scalar & dense parameter count, trainable generator parameter count, and compression ratio $P/\bar P$ \\
\bottomrule
\end{tabular}
\end{table}


\section{Tensorising DNN Layers}
\label{sec:Tensorization}

A prerequisite for replacing a classical DNN layer with an ADNTN decoder is a
well-defined, invertible mapping from the layer's parameter tensor to a
uniform high-order tensor on which the generator operates.  This
\emph{tensorisation} step is not merely a reshape: the choice of physical radix,
mode ordering, and padding strategy directly affects the expressivity and
compression efficiency of the generator.

Let $d\geq 2$ be the \emph{physical radix}, the common size of every mode of the
tensorised target.  For a layer with $P$ parameters we choose the smallest tensor
order $Q=\lceil \log_d P\rceil$ such that $d^Q\geq P$, pad the vectorised
parameters to length $d^Q$, and fold the result into an order-$Q$ tensor with
each mode of size $d$.  Throughout the paper we write
\[
  [d] := \{1,\ldots,d\},
\]
although zero-based indexing may be used in code.  The binary radix $d=2$ gives
the finest granularity and the highest nominal compression; larger values of $d$
reduce the tensor order $Q$ and may improve expressivity at the cost of higher
parameter counts in each core.

\subsection{Large Linear (Fully Connected) Layers}
Fully connected (FC) layers are the dominant parameter class in MLPs and in the
projection heads of transformer models, where a single weight matrix
$\mat{W}\in\R^{D_{\rm out}\times D_{\rm in}}$ may contain millions of entries.
With $P=D_{\rm out}D_{\rm in}$, let $\vecop(\mat{W})\in\R^P$ denote a fixed
column-major or row-major vectorisation.  The tensorised target is
\begin{equation}
  \ten{W}_{\rm tar}
  = \fold_{d,Q}\!\left(\operatorname{pad}_{d^Q}(\vecop(\mat{W}))\right)
  \in \R^{\underbrace{d\times\cdots\times d}_{Q\ \text{modes}}}.
  \label{eq:linear-tensorization}
\end{equation}
The inverse map removes the padding and reshapes the generated tensor back into
$\widehat{\mat{W}}_\Theta\in\R^{D_{\rm out}\times D_{\rm in}}$.

When the input and output dimensions have meaningful factors,
$D_{\rm out}\leq\prod_{q=1}^s O_q$ and
$D_{\rm in}\leq\prod_{q=1}^s I_q$, a paired tensorisation is often preferable:
\begin{equation}
  W_{o,i}
  \longleftrightarrow
  \ten{W}_{o_1,i_1,\ldots,o_s,i_s},
  \qquad
  o \leftrightarrow (o_1,\ldots,o_s),\quad
  i \leftrightarrow (i_1,\ldots,i_s).
  \label{eq:paired-tensorization}
\end{equation}
This ordering keeps local input-output interactions close in the tensor network
and it usually provides more efficient  compression  than a blind flattening.

\subsection{Large Convolutional Kernels}
Convolutional layers account for the majority of parameters and FLOPs in
standard vision models.  A 2D convolutional layer has a fourth-order kernel
\begin{equation}
  \ten{K}\in\R^{C_{\rm out}\times C_{\rm in}\times K_H\times K_W}.
\end{equation}
The total parameter count $P=C_{\rm out}C_{\rm in}K_HK_W$ can be large even for
small spatial filters when $C_{\rm out}$ and $C_{\rm in}$ are in the hundreds.
The simplest unified representation flattens all indices and then applies the
same fold as in Eq.~\eqref{eq:linear-tensorization}:
\begin{equation}
  \ten{K}
  \xrightarrow{\ \vecop\ }
  \vect{k}\in\R^{C_{\rm out}C_{\rm in}K_HK_W}
  \xrightarrow{\ \operatorname{pad},\ \fold_{d,Q}\ }
  \ten{W}_{\rm tar}\in\R^{d\times\cdots\times d}.
  \label{eq:conv-tensorization}
\end{equation}
For high accuracy it is often useful to preserve channel/spatial semantics in
the mode order: channel factors typically carry most of the parameter mass,
whereas $K_H$ and $K_W$ are small and should not be interleaved arbitrarily.

\subsection{Tensorising the Multi-Head Attention (MHA) Block}

In a Transformer attention block with embedding dimension $D$, the four dense
projection matrices are
\begin{equation}
  \mat{W}_Q,\ \mat{W}_K,\ \mat{W}_V,\ \mat{W}_O \in \R^{D\times D}.
\end{equation}
The three input projections can be compressed jointly by stacking them along the
output dimension into a single fused projection,
\begin{equation}
  \mat{W}_{\rm QKV} =
  \begin{bmatrix}\mat{W}_Q\\ \mat{W}_K\\ \mat{W}_V\end{bmatrix}
  \in \R^{3D\times D},
  \label{eq:qkv-concat}
\end{equation}
and then tensorising $\mat{W}_{\rm QKV}$ by Eq.~\eqref{eq:linear-tensorization}.
This joint tensorisation can expose shared structure among attention heads and
is consistent with recent tensorisation approaches to MHA compression
\citep{Gu2025}.  The output projection $\mat{W}_O$ can be compressed by a
separate generator or by a larger joint tensor if memory and optimisation budget
allow.

\section{Hierarchical Nonlinear Tensor Networks for DNN Compression}
\label{sec:architectures}

This section introduces the three nonlinear tensor-network decoders at the
heart of the ADNTN framework: the Tree Tensor Network (TTN), the augmented TTN
(aTTN), and the Multi-scale Entanglement Renormalisation Ansatz (MERA).  All
three share the same top-down decoding paradigm---a small latent root tensor is
expanded layer by layer through local core contractions into a high-order tensor,
which is finally reshaped into the target weight matrix or convolutional
kernel---but they differ in how information is mixed across branches during
expansion.

The TTN topology is loop-free (tree-structured): each internal bond separates
the tree into two independent subtrees.  This property makes TTNs simple to
batch and easy to train, but purely multilinear TTNs suffer from boundary
artefacts between branches that diverge early in the tree.  In numerical
multilinear algebra, a multilinear TTN is equivalent to a hierarchical
Tucker-type decomposition \citep{Shi2006,Cichocki2016}.  The \emph{nonlinear}
TTN studied here retains the same sparse tree topology but inserts a learnable
activation after each layer expansion, substantially enriching the family of
weight tensors that the generator can produce relative to its multilinear
counterpart.

\begin{figure}[H]
\centering
\incfig[0.8\textwidth]{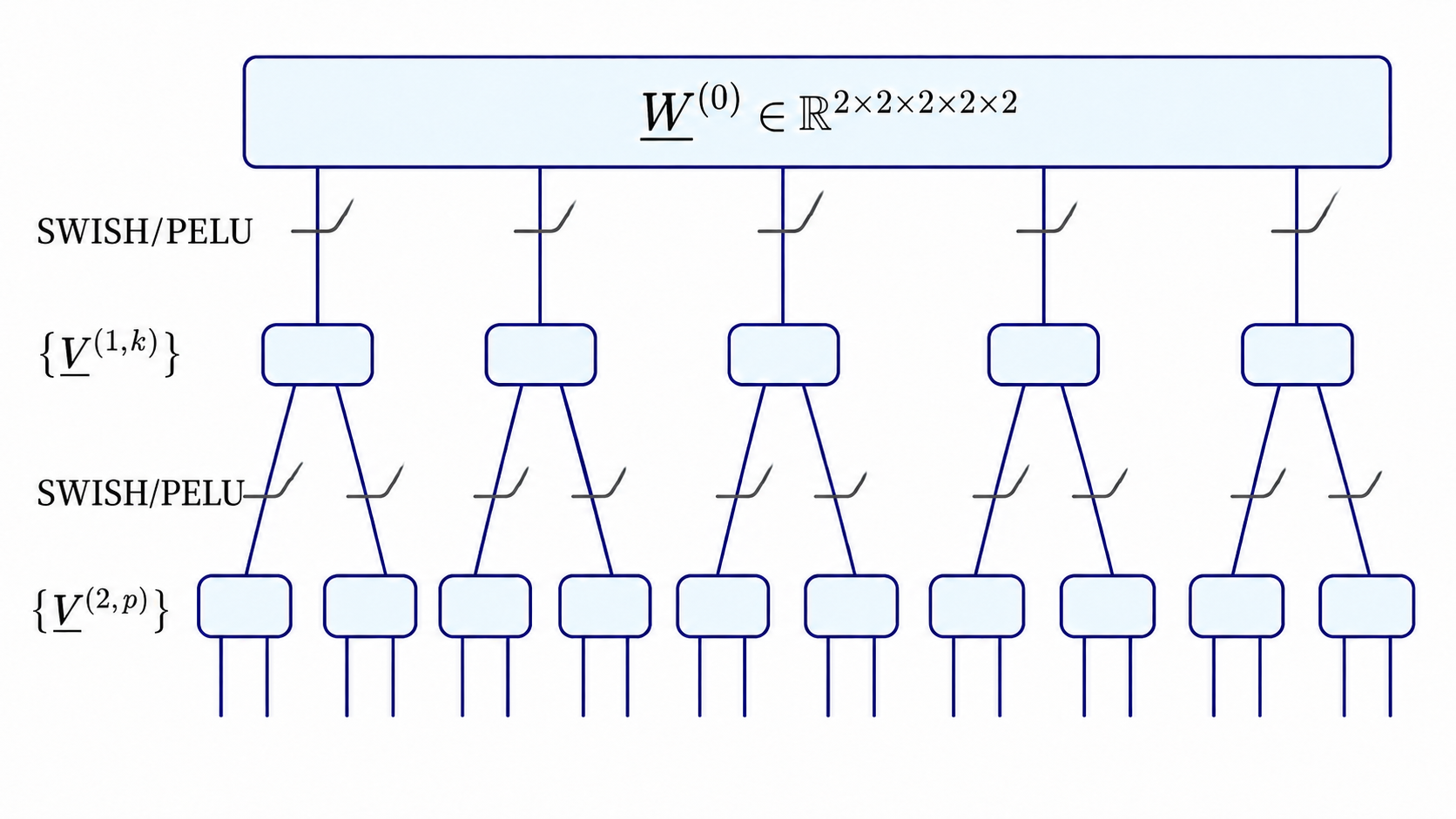}
\caption{\textbf{Nonlinear Tree Tensor Network (TTN) decoder used as a weight
generator.}  A small latent tensor $\ten{W}^{(0)}$ of bottleneck order
$Q_{\rm bot}=5$ is expanded layer by layer by rank-3 cores
$\ten{V}^{(\ell,k)}$.  Learnable nonlinear activations (SWISH or PELU by default) are applied after each hidden expansion layer, while the output layer is kept linear so that the generated tensor can be directly reshaped into an ordinary linear or convolutional weight.}
\label{fig:TTN}
\end{figure}

The aTTN augments a TTN by inserting one or more layers of local two-site
\emph{disentanglers} near the physical (output) boundary of the generator.
These lateral tensors mix neighbouring branches before the final weight tensor is
read out, directly counteracting the cross-branch discontinuities that appear
when distant leaves share no common contraction path in a strict tree.  The cost
of adding boundary disentanglers is modest: the number of extra parameters grows
only linearly in the number of physical modes.  MERA pushes this idea to its
logical conclusion by inserting disentanglers at \emph{every} scale of the
hierarchy, creating overlapping causal cones and the strongest possible
multi-scale information mixing \citep{Vidal2007,Evenbly2014}.  The price is
higher memory and permutation overhead, but the expressive gain is significant
for layers whose weight structure exhibits correlations at multiple scales.

\begin{figure}[H]
\centering
\incfig[0.48\textwidth]{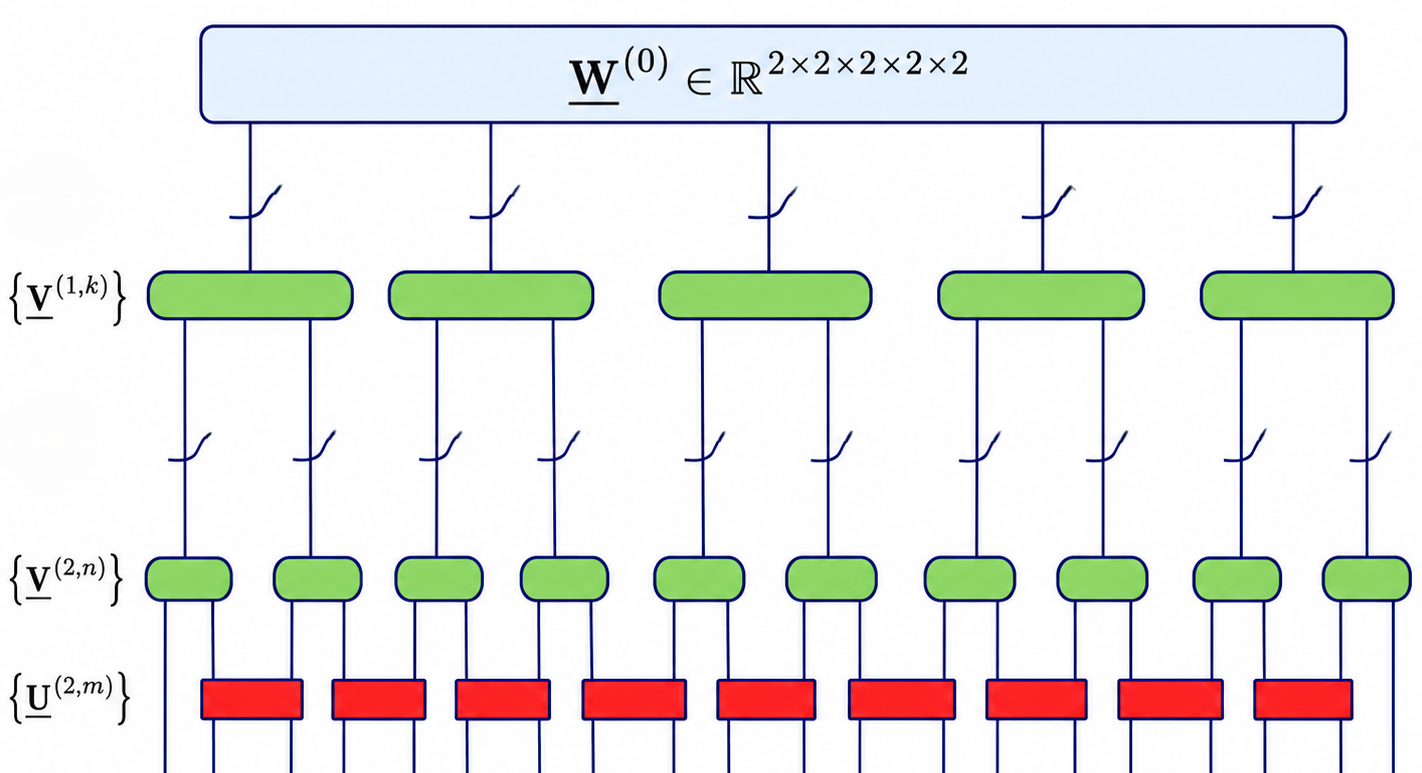}\hfill
\incfig[0.48\textwidth]{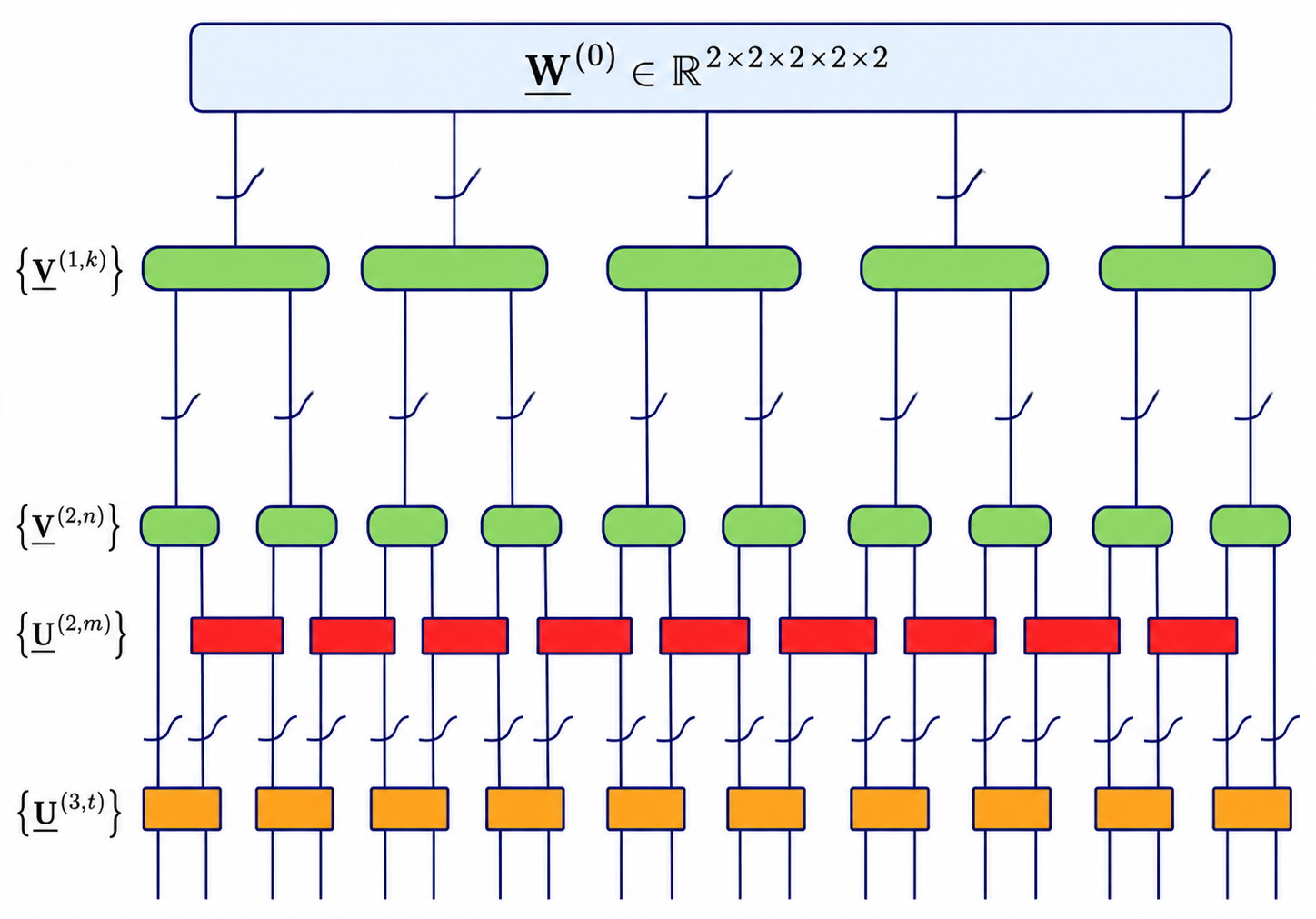}
\caption{\textbf{Augmented TTN (aTTN) decoders with one (left) and two (right)
boundary-disentangler layers.}  The TTN expansion cores (green cores) build the
hierarchical tensor, while the two-site cores (red/orange) mix neighbouring
physical branches before readout.  These lateral tensors reduce boundary
artefacts at modest parameter cost; unlike quantum MERA, they are trained as
unconstrained real-valued parameters unless a soft isometry regulariser
(Eq.~\eqref{eq:isometry-penalty}) is added.}
\label{fig:aTTN}
\end{figure}

\begin{figure}[H]
\centering
\incfig[0.8\textwidth]{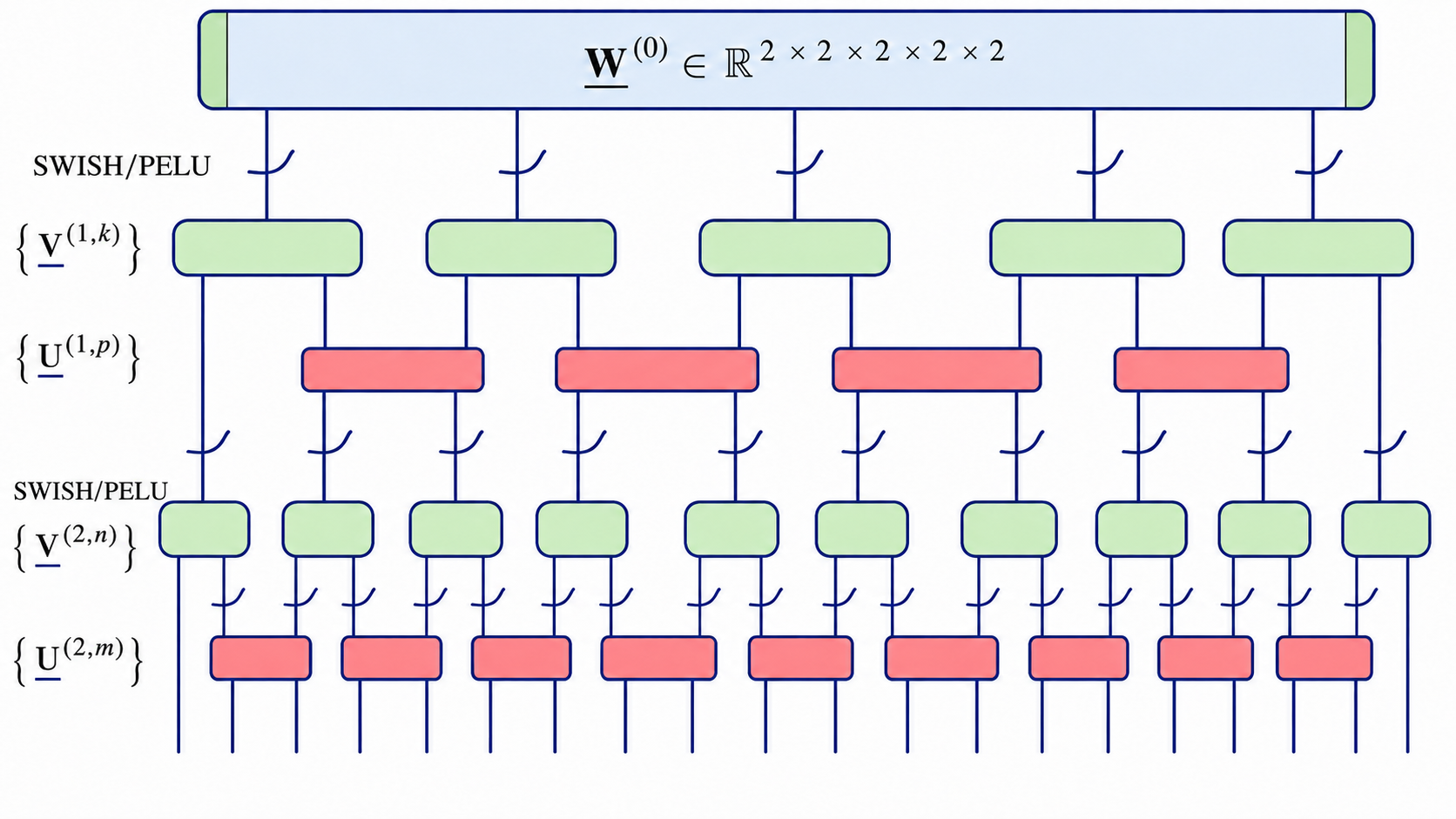}
\caption{\textbf{Nonlinear MERA-style decoder for ADNTN weight generation.}  Each
scale alternates vertical expansion cores $\ten{V}^{(\ell,k)}$ with lateral
disentanglers $\ten{U}^{(\ell,k)}$, producing overlapping causal cones and
stronger multi-scale mixing than a strict tree.  The diagram shows the
open-boundary version used in this paper, with learnable nonlinear activations (SWISH or PELU) placed between complete generator layers.}
\label{fig:MERA}
\end{figure}

\subsection{Quantised Local Cores Nonlinearly Interconnected}
\label{sec:quant}

A central design principle in ADNTN compression is to keep all local dimensions
small.  For a homogeneous binary architecture we often use physical radix
$d=2$ and bond dimension $\chi=2$.  The canonical core shapes are
\begin{align}
  \text{chain/branching core:}\qquad
  &\ten{G}\in\R^{\chi\times d\times\chi},
  \label{eq:core-chain}\\
  \text{binary expansion core:}\qquad
  &\ten{V}\in\R^{\chi\times\chi\times\chi},
  \label{eq:core-expansion}\\
  \text{two-site disentangler:}\qquad
  &\ten{U}\in\R^{\chi\times\chi\times\chi\times\chi},
  \label{eq:core-disentangler}\\
  \text{leaf map:}\qquad
  &\ten{P}\in\R^{\chi\times d}.
  \label{eq:core-leaf}
\end{align}
For a binary tree with $Q$ leaves and bounded $\chi$, a representative
parameter count is
\begin{equation}
  \bar P_{\rm TTN}
  \approx (Q-1)\chi^3 + Q\chi d,
  \label{eq:ttn-count}
\end{equation}
whereas a dense order-$Q$ target tensor has $d^Q$ entries.  Thus, for fixed $d$ and $\chi$, the nominal parameter compression grows
\emph{exponentially} with the target tensor order $Q$---this is the fundamental
advantage of hierarchical tensor-network generators over flat low-rank methods.
For a binary architecture with $\chi=2$ and a target layer of order $Q=30$,
the generator has only a few thousand parameters while the target tensor has over
$10^9$ entries.  This is a parameter-count statement; accuracy, latency, and
energy gains must still be verified empirically and depend on the contraction
schedule and hardware target.  Nevertheless, the exponential scaling is a
mathematically rigorous property of the ADNTN architecture and is the primary
motivation for investigating it as a compression strategy.

\begin{remark}[Matching the generated order to the target]
\label{rem:order-matching}
Equation~\eqref{eq:ttn-count} describes the homogeneous binary case, in which the
generator has $Q$ physical legs.  In general the generator is read \emph{top
down}: a latent root of order $Q_{\rm bot}$ is expanded layer by layer, and the
per-layer branching factor and radix are chosen so that the final generated
tensor matches the padded target order of Eq.~\eqref{eq:linear-tensorization}.
The pure binary doubling schedule $N_\ell = 2N_{\ell-1}$ is therefore only a
special case; when $2^{L}Q_{\rm bot}$ does not coincide with the required target
order, the architecture reconciles the two using a non-binary final expansion, a
mixed radix across layers, padding, or row grouping (Section~\ref{sec:grouping}).
The reported parameter counts $\bar P$ in Section~\ref{sec:experiments} count all
latent, expansion, disentangler, and leaf parameters of the actual schedule used,
not the idealised count of Eq.~\eqref{eq:ttn-count}.
\end{remark}

The binary radix $d=2$ is attractive because it gives extreme compression,
uniform tensor legs, and power-of-two memory layouts.  Larger $d$ or $\chi$ can
be introduced adaptively when validation loss indicates that the bottleneck is
too restrictive.

\subsection{Practical Architectural Choices}
\label{sec:practical}

\textbf{Mode ordering.}
Mode ordering is a modelling choice with a direct impact on compression
quality---it is emphatically not a cosmetic detail.  The ADNTN generator
approximates the target tensor by exploiting locality in the chosen mode
ordering: modes that are adjacent in the tree contract early and share bond
capacity, while modes that are far apart in the tree communicate only through
the narrow bottleneck at the root.  For fully connected layers, paired indices
$(o_q,i_q)$ interleave output and input modes and preserve local input-output
structure, typically yielding better compression than a blind flattening.  For
convolutional kernels, the spatial modes $K_H$ and $K_W$ are small and should
\emph{not} be interleaved arbitrarily with the large channel modes $C_{\rm out}$
and $C_{\rm in}$, which dominate the parameter count.  Suboptimal mode ordering
hides low-dimensional structure from the generator, forces unnecessary growth in
$\chi$ to compensate, and can substantially increase the total parameter count
$\bar P$ without improving accuracy.

\subsection{Topology Comparison: TTN, aTTN, and MERA}
\label{sec:topology_comparison}

\begin{table}[H]
\centering
\caption{\textbf{Qualitative comparison of the three ADNTN topologies studied in
this paper.}  TTN is the lowest-cost baseline, aTTN adds boundary mixing to
reduce tree artefacts, and MERA applies mixing at every scale for greater
expressivity at higher implementation cost.}
\label{tab:topology}
\begin{tabular}{p{0.16\textwidth}p{0.36\textwidth}p{0.36\textwidth}}
\toprule
\textbf{Topology} & \textbf{Advantages} & \textbf{Limitations} \\
\midrule
TTN
  & Simple $O(\log Q)$ depth, low parameter count, and strong batching potential.
  & Possible boundary artefacts and weaker interactions across early tree splits. \\
\addlinespace
aTTN
  & Boundary disentanglers improve lateral information flow at modest cost.
  & Extra permutations and boundary contractions add implementation complexity. \\
\addlinespace
MERA
  & Overlapping light cones and multi-scale mixing; often more expressive.
  & Larger constants in memory, tensor permutation, and contraction scheduling. \\
\bottomrule
\end{tabular}
\end{table}

A practical design path is to begin with TTN, move to aTTN when generated
weights show block or cross-branch artefacts, and use MERA when multi-scale or
long-range mixing is essential and the additional contraction overhead is
acceptable.  This incremental strategy also facilitates ablation studies that
isolate the contribution of each architectural component.

\subsection{Comparison with the Brick-Wall ADTN}
\label{sec:brickwall_comparison}

Because the brick-wall ADTN of \citet{Qing2025} is the most direct predecessor of
this work, it is worth stating precisely how the hierarchical generators differ
from it.  In a brick-wall ADTN, the layer parameters are encoded into the
contraction
\begin{equation}
  \mathcal{T}
  = \mat{L}^{(M)}\sigma\!\Bigl(\cdots
      \mat{L}^{(2)}\sigma\bigl(\mat{L}^{(1)}\,\textstyle\bigotimes_{q=1}^{Q}\vect{v}\bigr)\Bigr),
  \label{eq:brickwall-map}
\end{equation}
where each effective layer matrix $\mat{L}^{(m)}$ is built from a column of
four-index cores $\ten{A}^{[k]}\in\R^{d\times d\times d\times d}$, $\vect{v}$ is a
fixed boundary vector, and $\sigma$ is a \emph{fixed} ReLU.  Information travels
one column at a time, so two physical indices separated by $\Delta$ positions can
interact only after at least $\Delta$ tensor layers.  Coupling the two ends of a
large layer therefore requires depth $M=O(Q)$, and each additional layer adds
$O(Q)$ parameters while lengthening the gradient path.

The hierarchical generators studied here change three things, summarised in
Table~\ref{tab:brickwall}.  First, \emph{connectivity}: a TTN/aTTN/MERA decoder
couples all $Q$ modes through a logarithmic-depth tree, so global information flow
costs only $O(\log Q)$ layers instead of $O(Q)$; long-range correlations are
mediated by the narrow root bottleneck rather than by a long chain of local
bricks.  Second, \emph{nonlinearity}: the fixed ReLU is replaced by learnable,
self-gated activations (Section~\ref{sec:learnable-nonlin}) that remain
differentiable on the negative branch and so keep gradients alive through the
$\chi=2$ bonds.  Third, \emph{lateral mixing}: aTTN adds boundary disentanglers
and MERA adds disentanglers at every scale, mechanisms with no analogue in a
strict brick wall, which directly repair the cross-branch discontinuities that a
loop-free tree would otherwise exhibit.  Finally, training is simplified: instead
of the two-stage Frobenius-then-task procedure of \citet{Qing2025}, all ADNTN
topologies are trained end-to-end from the single initialisation policy of
Section~\ref{sec:init}.

\begin{table}[H]
\centering
\caption{\textbf{Brick-wall ADTN \citep{Qing2025} versus the hierarchical ADNTN
generators of this paper.}  The two families differ in how information is routed
across the generation graph, in the nonlinearity, in the presence of lateral
mixing, and in the training recipe.}
\label{tab:brickwall}
\begin{tabular}{p{0.30\textwidth}p{0.30\textwidth}p{0.30\textwidth}}
\toprule
\textbf{Property} & \textbf{Brick-wall ADTN} & \textbf{Hierarchical ADNTN (this paper)} \\
\midrule
Generation-graph topology
  & Flat brick wall of four-index cores.
  & Logarithmic-depth tree (TTN/aTTN/MERA). \\
\addlinespace
Depth to couple distant modes
  & $O(Q)$ tensor layers.
  & $O(\log Q)$ generator layers. \\
\addlinespace
Nonlinearity
  & Fixed ReLU between layers.
  & Learnable self-gated activations (SWISH/SiLU, PELU, ELU/GELU, SwiGLU). \\
\addlinespace
Lateral mixing
  & None.
  & Boundary disentanglers (aTTN); multi-scale disentanglers (MERA). \\
\addlinespace
Training
  & Two-stage: Frobenius pre-training, then task loss.
  & End-to-end from a single global initialisation policy. \\
\bottomrule
\end{tabular}
\end{table}

\subsection{Dynamic Rank and Radix Growth}
\label{sec:rank_growth}

Starting with the minimum radix $d=2$ and bond dimension $\chi=2$ maximises the
compression ratio, but some layers may have insufficient generator capacity at
this setting, as evidenced by high reconstruction loss or degraded task accuracy
relative to the dense baseline.  Rather than committing to a large $\chi$ from
the outset---which would inflate the parameter count and reduce compression---we
advocate a \emph{progressive growth} strategy: begin with the smallest viable
$(d,\chi)$, train until convergence, and expand individual cores only where
the optimiser signals insufficient capacity.  A simple growth step for an
expansion core is
\begin{equation}
  \ten{V}\in\R^{\chi\times\chi\times\chi}
  \longrightarrow
  \ten{V}'\in\R^{(\chi+\Delta\chi)\times(\chi+\Delta\chi)\times(\chi+\Delta\chi)},
  \label{eq:rank-growth}
\end{equation}
where the leading $\chi^3$ block of $\ten{V}'$ is initialised with the learned
entries of $\ten{V}$ and the remaining entries are initialised near zero.  This
warm start approximately preserves the learned generating function while
providing the optimiser with additional degrees of freedom precisely where the
current generator is capacity-limited.  The growth decision can be guided by
gradient norms, held-out reconstruction error, or a saliency score computed on
the boundary of the core.

\subsection{Initialisation}
\label{sec:init}

Proper initialisation is one of the most practically important, yet most
underspecified, aspects of training deep tensor-network generators.  Unlike
shallow networks, where standard He or Xavier rules apply almost universally,
ADNTNs stack many small contraction layers, each contributing multiplicatively to
the signal magnitude.  A signal travels through $L$ expansion layers before it
reaches the generated weight tensor, so the effective variance of the generated
weight scales as the product of core variances across all layers.  If the core
variances are too large, intermediate tensors explode or drive the nonlinearities
into saturation; if they are too small, the adjoint signal vanishes before
reaching the latent root tensor.  For this reason we adopt a \emph{single,
globally consistent initialisation policy} across TTN, aTTN, and MERA, rather
than mixing He Normal, Xavier Uniform, and ad-hoc Gaussian rules across
different experiments---a practice that has led to irreproducible results in
prior tensor-network compression work.

For a core $\ten{C}$, define its effective fan-in and fan-out by grouping the
legs that are contracted with the incoming state and the legs that are exposed
to the outgoing state:
\begin{equation}
  n_{\rm in}(\ten{C})=\prod_{q\in I_{\rm in}(\ten{C})} n_q,
  \qquad
  n_{\rm out}(\ten{C})=\prod_{q\in I_{\rm out}(\ten{C})} n_q,
  \label{eq:fanin-fanout}
\end{equation}
where $n_q$ is the size of leg $q$.  For the homogeneous binary expansion core
$V_{p,a,b}$ this gives $n_{\rm in}=d$ and $n_{\rm out}=d^2$; for a two-site
mixing core $U_{\alpha,\beta,a,b}$ it gives
$n_{\rm in}=n_{\rm out}=d^2$.  If physical and virtual dimensions differ, the
same rule is applied using the actual leg sizes.

For hidden cores followed by SWISH,  PELU,  or SwiGLU-type
activations, the default is a He/Kaiming-style normal initialisation
\citep{He2015}:
\begin{equation}
  C_i \sim \mathcal{N}\!\left(0,\frac{g_\sigma^2}{n_{\rm in}(\ten{C})}\right),
  \qquad
  g_\sigma \approx \sqrt{2}
  \quad \text{for ReLU-family activations}.
  \label{eq:he-core-init}
\end{equation}
For PReLU with negative slope $a$, a useful gain is
$g_\sigma=\sqrt{2/(1+a^2)}$ \citep{He2015}.  For final linear readout cores,
for approximately linear activations, or when the next activation is close to
odd and variance preserving near the origin, we use Xavier/Glorot scaling
\citep{GlorotBengio2010}:
\begin{equation}
  C_i \sim \mathcal{U}\!\left[-g_\sigma\sqrt{\frac{6}{n_{\rm in}+n_{\rm out}}},
            \;g_\sigma\sqrt{\frac{6}{n_{\rm in}+n_{\rm out}}}\right].
  \label{eq:xavier-core-init}
\end{equation}
SIREN-style periodic activations require their own scaling: the first layer and
subsequent layers should follow the frequency-dependent rules in
\citet{Sitzmann2020}, otherwise the sine generator can become either nearly
constant or chaotic.

The latent root is initialised separately,
\begin{equation}
  W^{(0)}_{i_1\ldots i_{Q_{\rm bot}}}
  \sim \mathcal{N}(0,s_0^2),
  \label{eq:latent-init}
\end{equation}
with $s_0$ chosen so that the first generated weights have a reasonable scale.
In the simulations in Section~\ref{sec:experiments} we use $s_0=0.05$ unless a
teacher-based warm start is available.  Disentanglers are preferably initialised
near the identity map,
\begin{equation}
  U_{\alpha,\beta,a,b}
  = \delta_{\alpha a}\delta_{\beta b}+\tau R_{\alpha,\beta,a,b},
  \qquad
  R_{\alpha,\beta,a,b}\sim\mathcal{N}\!\left(0,\frac{g_\sigma^2}{n_{\rm in}}\right),
  \label{eq:identity-disentangler-init}
\end{equation}
with a small $\tau$ such as $10^{-3}$--$10^{-1}$.  This allows aTTN and MERA to
start close to the corresponding TTN while leaving AD free to learn lateral
mixing.

In practice, the following reproducible policy is recommended: use
Eq.~\eqref{eq:he-core-init} for hidden nonlinear expansion cores,
Eq.~\eqref{eq:xavier-core-init} for final linear readout cores,
Eq.~\eqref{eq:identity-disentangler-init} for lateral disentanglers, and
Eq.~\eqref{eq:latent-init} for the latent root.  After the first forward pass,
a lightweight calibration step can rescale each hidden layer so that
$\Var[\ten{Z}^{(\ell)}]$ remains close to a chosen target, typically one.  When
a dense teacher weight is available and small enough to tensorise, TT-SVD or
Tucker initialisation followed by nonlinear fine-tuning can further accelerate
convergence \citep{Oseledets2011,KoldaBader2009}.

\subsection{Physics Tensor Networks versus Machine-Learning Generators}
\label{sec:physics_vs_ml}

Although ADNTNs borrow their diagrammatic language and topological vocabulary
from quantum many-body physics, the objectives and constraints are fundamentally
different.  In quantum physics, tensor networks such as MERA represent
normalised many-body wavefunctions and core tensors are constrained to be
isometric or unitary to preserve norm and entanglement structure.  In
neural-network compression, the generated object is an unconstrained real-valued
weight tensor whose singular value spectrum, scale, and sign are determined
entirely by the downstream task loss.  This distinction matters both
theoretically and practically.  For real-valued cores, the physics isometry constraints are
\begin{equation}
  \sum_{a,b} V_{i,a,b}V_{j,a,b}=\delta_{ij},
  \qquad
  \sum_{a,b} U_{a,b,i,j}U_{a,b,i',j'}=\delta_{ii'}\delta_{jj'}.
  \label{eq:isometry}
\end{equation}
In neural-network compression, the generated object is an unconstrained
real-valued weight tensor.  Strict isometry/unitarity is therefore not imposed
by default: neural weights may need arbitrary scale, sign, and anisotropic
spectra.  However, soft penalties such as
\begin{equation}
  \loss_{\rm iso}
  = \lambda_V\sum_{\ell,k}\left\|\mat{V}_{(i),(ab)}^{(\ell,k)}
       \mat{V}_{(i),(ab)}^{(\ell,k)\,T}-\mat{I}\right\|_F^2
    + \lambda_U\sum_{\ell,k}\left\|\mat{U}_{(ab),(ij)}^{(\ell,k)}
       \mat{U}_{(ab),(ij)}^{(\ell,k)\,T}-\mat{I}\right\|_F^2
  \label{eq:isometry-penalty}
\end{equation}
can be useful regularisers when numerical stability or bounded activations are
important.

DMRG/TEBD-type methods optimise one core at a time by constructing environment
tensors and performing local solves or SVD sweeps.  These methods are powerful
for multilinear physics models, but nonlinear activations and end-to-end task
losses break the local linear subproblem.  AD is therefore more natural for
ADNTNs: it differentiates the complete executed program, including nonlinearities,
loss terms, batching, and any downstream network layers \citep{Liao2019,Qing2025}.

\section{Why Use a Nonlinear Generator for a Linear Layer?}
\label{sec:nonlinearity}

There is no contradiction in using a nonlinear generator for a layer that is
linear at inference time.  The ADNTN construction separates two computational
graphs.
\begin{itemize}[leftmargin=*]
  \item The \textbf{execution graph} maps data to predictions, for example
        $\vect{y}=\widehat{\mat{W}}_\Theta\vect{x}$ or
        $\ten{Y}=\widehat{\ten{K}}_\Theta\star\ten{X}$.  This graph may remain
        linear in the input.
  \item The \textbf{generation graph} maps a latent tensor and core tensors to
        weights,
        $\widehat{\ten{W}}_\Theta=\mathcal{G}_\Theta(\ten{W}^{(0)})$.  This map
        may be nonlinear in the cores and intermediate states.
\end{itemize}
Thus the generated operator can be used as an ordinary linear layer even when
its parameterisation is nonlinear.

\textbf{Relaxing multilinear rank restrictions.}
For a purely multilinear tensor network, any bipartition $A|B$ of the physical
indices satisfies a cut-rank bound
\begin{equation}
  \rank\!\left(\unfold_{A|B}(\ten{W})\right)
  \leq \prod_{e\in\partial(A,B)}\chi_e
  \leq \chi^{|\partial(A,B)|},
  \label{eq:cut-rank-bound}
\end{equation}
where $\partial(A,B)$ is the set of virtual bonds cut by the bipartition.  For a
TT/MPS cut, $|\partial(A,B)|=1$, giving the familiar rank bound $\leq\chi$.
For TTN and MERA the bound depends on the chosen cut and topology, but the key
point remains: a linear tensor network has algebraic ranks controlled by its
virtual bonds.

Elementwise nonlinearities break the multilinear factorisation.  A layer of the
form
\begin{equation}
  \ten{W}^{(\ell)}
  = \phi_\ell\!\left(\Contract_\ell\!\bigl(\ten{W}^{(\ell-1)},
       \{\ten{C}^{(\ell,k)}\}_k\bigr)\right)
  \label{eq:generic-nonlinear-layer}
\end{equation}
is no longer multilinear in the cores.  Even in the matrix case, if
$\mat{M}=\mat{A}\mat{B}$ has $\rank(\mat{M})\leq\chi$, the matrix
$\phi(\mat{M})$ obtained by applying a nonlinear function entrywise can be full
rank for generic $\mat{A}$ and $\mat{B}$.  Consequently, the fixed-cut rank
bound in Eq.~\eqref{eq:cut-rank-bound} no longer directly characterises the
final generated weight matrix.

\begin{remark}
Nonlinear TN generators remove the strict multilinear rank bottleneck and allow the model to
learn a richer low-dimensional manifold of weights.  The optimistic empirical
hypothesis, supported by compression results, is that many trained DNN layers lie
close enough to such structured nonlinear manifolds to be approximated with very
few trainable degrees of freedom.
\end{remark}

\textbf{Gradient flow.}
The activation should preserve useful derivatives along the deep generator.  A
standard ReLU has zero derivative for negative pre-activations, which can freeze
paths through a small-bond tree.  Smooth activations with nonzero or learnable
negative-side behaviour are therefore preferable in tensor networks with  narrow narrow-bond regime ($d=\chi=2$).

\subsection{Learnable Nonlinearities for Enhanced Expressivity}
\label{sec:learnable-nonlin}

A defining feature of ADNTNs, and a key point of departure from the fixed-ReLU
brick-wall ADTN of \citet{Qing2025}, is that the activations interleaved between
contraction layers are themselves \emph{trainable}.  In the narrow-bond regime
($d=\chi=2$) that gives the highest compression, the choice and shape of the
activation strongly affects both expressivity and gradient flow.  We therefore
treat the nonlinearity as part of the learnable model rather than as a fixed
hyperparameter.  Below we summarise the fixed and adaptive activations evaluated
in this work.

\paragraph{Default fixed activations: GELU and ELU.}
Our default fixed choices are GELU and ELU,
\begin{align}
  \operatorname{GELU}(x)
  &= x\,\Phi(x)
   = \frac{x}{2}\left(1+\operatorname{erf}\!\left(\frac{x}{\sqrt{2}}\right)\right),
  \label{eq:gelu}\\
  \operatorname{ELU}_\alpha(x)
  &= \begin{cases}
      x, & x>0,\\
      \alpha(e^x-1), & x\leq 0.
     \end{cases}
  \label{eq:elu}
\end{align}
GELU weights its input by the Gaussian cumulative probability $\Phi(x)$
\citep{Hendrycks2016}, while ELU shifts negative activations below zero and
improves mean-activation dynamics \citep{Clevert2015}.  Unlike ReLU, both are
smooth and have nonzero derivatives for $x<0$, which prevents the dead-path
problem that fixed ReLU can cause in $\chi=2$ generators.

\paragraph{Adaptive self-gating: SWISH/SiLU.}
The Swish/SiLU activation,
\begin{equation}
  f(x)=x\,\operatorname{sigmoid}(\beta x)
  =\frac{x}{1+e^{-\beta x}},
  \label{eq:swish}
\end{equation}
is a smooth, non-monotone self-gating function \citep{Ramachandran2017}.  At
$\beta=0$ it collapses to the linear map $f(x)=x/2$.  The gate parameter $\beta$
controls the depth of the non-monotone ``negative valley'': a small $\beta$
stretches the curve towards linearity, yielding smooth gradients that mitigate
vanishing gradients during the early, volatile phase of training, whereas a large
$\beta$ drives the function towards a ReLU shape.  Treating $\beta$ as a learnable
parameter lets different parts of the tensor network choose dynamically between
near-linear smoothing and hard thresholding, rather than imposing a single fixed
nonlinearity on every contraction.

\paragraph{Parametric exponential units: PELU and MPELU.}
PELU (Parametric Exponential Linear Unit) augments ELU with learnable parameters,
\begin{equation}
f(x)=
  \begin{cases}
    \beta x, & x > 0,\\[2pt]
    \gamma\,\alpha\!\left(\exp\!\left(\tfrac{x}{\alpha}\right)-1\right), & x \le 0,
  \end{cases}
\label{eq:pelu}
\end{equation}
where $\alpha>0$ sets the scale of the negative exponential branch, $\beta>0$ the
slope of the positive linear branch, and $\gamma>0$ the magnitude of the negative
values.  Unlike ReLU, PELU is continuous and smooth at the origin, giving smoother
gradients during backpropagation. 

\textbf{MPELU (Multiple Parametric Exponential Linear
Unit):} acts as a unified generalisation layer that collapses ReLU, LReLU, PReLU,
and ELU into a single framework using channel- or layer-specific parameters,
\begin{equation}
f(x_c)=
  \begin{cases}
    x_c, & x_c>0,\\[2pt]
    \alpha_c\!\left(\exp(\beta_c x_c)-1\right), & x_c\le 0,
  \end{cases}
\label{eq:mpelu}
\end{equation}
where $c$ indexes the channel or layer unit.  Depending on the learned
$(\alpha_c,\beta_c)$, MPELU recovers ELU ($\alpha_c=\beta_c=1$), ReLU
($\alpha_c=0$), or PReLU (small $\beta_c$, e.g.\ $0.01$).  In ADNTNs, such
adaptive activations let AD tune the nonlinearity per layer and so optimise the
reconstruction--task trade-off directly.

\begin{figure}[H]
\centering
\incfig[0.48\textwidth]{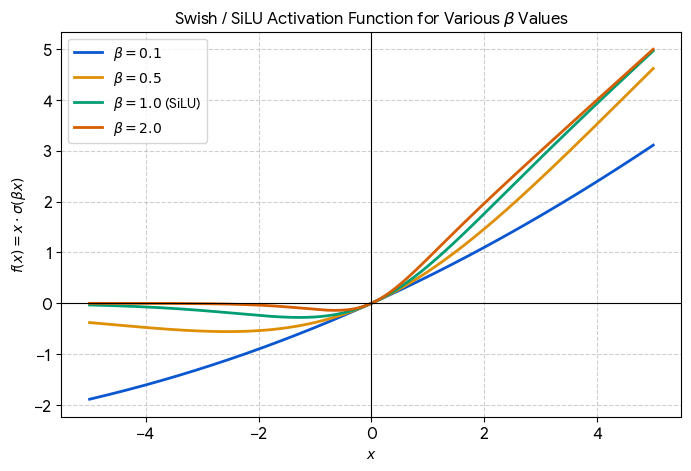}\hfill
\incfig[0.48\textwidth]{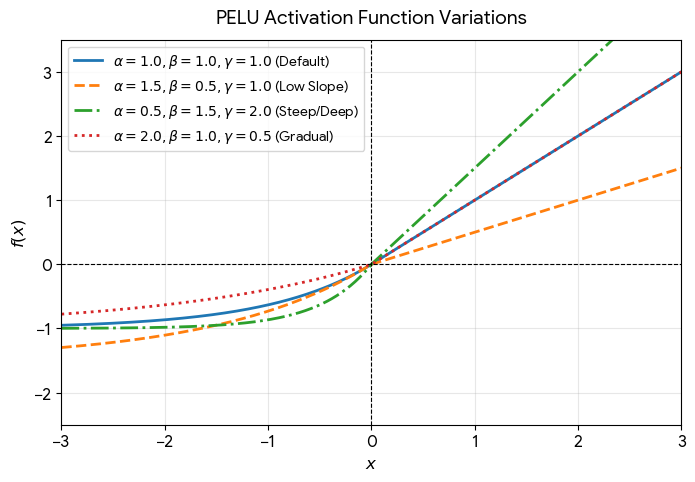}
\caption{Adaptive (learnable) SWISH/SiLU (left) and PELU (right) for several
parameter settings.  As the gate or scale parameters vary, both families
interpolate continuously between near-linear and strongly rectifying behaviour,
giving AD a smooth landscape over which to optimise the per-layer nonlinearity.}
\label{fig:activations}
\end{figure}

\paragraph{Further options.}
Three additional families are promising for nonlinear core tensor contractions:

\textbf{Periodic (SIREN-style) activations:}
\begin{equation}
  f(x)=\sin(\omega_0 x),
  \label{eq:siren-activation}
\end{equation}
require careful initialisation but can represent oscillatory or sharp local
structure \citep{Sitzmann2020}. 

 \textbf{Learnable rational activations},
\begin{equation}
  f(x)=\frac{P_m(x)}{Q_n(x)}
  =\frac{a_0+a_1x+\cdots+a_mx^m}{1+b_1x+\cdots+b_nx^n},
  \label{eq:rational-activation}
\end{equation}
have trainable coefficients and can approximate smooth functions efficiently,
providing a flexible activation family \citep{Boulle2020}.

\textbf{SwiGLU (Swish Gated Linear Unit)} activation function belongs to the family of Gated Linear Units (GLU). Introduced by  \citep{SwiGLU}, 
it incorporates an input-dependent gating mechanism that scales information dynamically. 

For a given input vector $\bm{x} \in \mathbb{R}^d$, the standard SwiGLU layer is defined using two distinct linear projections via weight matrices $\mat W$ and $\mat V$:
\begin{equation}
f(\bm{x}) =\text{SwiGLU}(\bm{x}) = \text{Swish}_1(\mat{W}\bm{x}) \odot  \mat{V}\bm{x} 
\end{equation}
where $\odot$ denotes the element-wise (Hadamard) product, and $\text{Swish}_1(z)$ (or SiLU) is the point-wise non-monotonic activation function:
\begin{equation}
\text{Swish}_1(z) = z \cdot \sigma(\beta z) = \frac{z}{1 + e^{-\beta z}}
\end{equation}

In a standard linear tensor network, core contractions act as multi-linear maps. Replacing standard point-wise activations like ReLU with SwiGLU requires mapping the contraction into parallel \textbf{Gate} and \textbf{Value} paths.

Let $\mathcal{A}$ and $\mathcal{B}$ be two core tensors slated for contraction over a set of shared modes. To apply SwiGLU, we instantiate two structurally identical but independent weight variants of the target tensor: $\mathcal{B}^{(W)}$ (the Gate tensor) and $\mathcal{B}^{(V)}$ (the Value tensor). 

The general procedure follows three steps:
\begin{enumerate}
    \item \textbf{Gate Path Contraction:} Compute intermediate tensor $\mathcal{G} = \text{Contract}(\mathcal{A}, \mathcal{B}^{(W)})$.
    \item \textbf{Value Path Contraction:} Compute intermediate tensor $\mathcal{V} = \text{Contract}(\mathcal{A}, \mathcal{B}^{(V)})$.
    \item \textbf{Element-wise Gating:} Compute the final non-linear output tensor $\mathcal{Y} = \text{Swish}(\mathcal{G}) \odot \mathcal{V}$, where $\odot$ is the Hadamard product.
\end{enumerate}
\section{General Loss Functions for AD Training}
\label{sec:loss}

ADNTNs are not restricted to a Frobenius reconstruction objective.  Since the
generator is a differentiable program, any differentiable or subdifferentiable
scalar loss can be used, including reconstruction, cross-entropy, distillation,
regularisation, quantisation-aware, and hardware-aware terms.

Historically, tensor-network compression matched a pre-trained target tensor by
minimising
\begin{equation}
  \loss_{\rm Frob}(\Theta)
  = \frac{1}{2}\left\|\widehat{\ten{W}}_\Theta-\ten{W}_{\rm tar}\right\|_F^2.
  \label{eq:frobenius-loss}
\end{equation}
This objective is useful for layer-wise fitting, but it treats all parameter
entries as equally important.  In classification, the task loss depends on the
logits, not directly on entrywise reconstruction error.

For a mini-batch $\{(\vect{x}_b,\vect{y}_b)\}_{b=1}^B$ with logits
$\vect{z}_b=F(\vect{x}_b;\widehat{\ten{W}}_\Theta)$ and predicted probabilities
$\widehat{\vect{y}}_b=\softmax(\vect{z}_b)$, the cross-entropy loss is
\begin{equation}
  \loss_{\rm CE}(\Theta)
  = -\frac{1}{B}\sum_{b=1}^{B}\sum_c y_{b,c}\log \widehat y_{b,c}.
  \label{eq:ce-loss}
\end{equation}
The adjoint entering the generator is obtained by the vector-Jacobian product
\begin{equation}
  \overline{\ten{W}}
  := \frac{\partial \loss_{\rm CE}}{\partial \widehat{\ten{W}}_\Theta}
  = \frac{1}{B}\sum_{b=1}^{B}
  \left(\frac{\partial \vect{z}_b}{\partial \widehat{\ten{W}}_\Theta}\right)^{\!*}
  \left(\widehat{\vect{y}}_b-\vect{y}_b\right),
  \label{eq:adjoint-ce-correct}
\end{equation}
where $^*$ denotes the adjoint (transpose for real tensors) of the local
Jacobian map.  This expression is exactly what reverse-mode AD computes without
forming the Jacobian explicitly.

A general nonnegative mixture is
\begin{equation}
  \loss(\Theta)
  = \sum_{m=1}^{M}\lambda_m\loss_m(\Theta)
    + \lambda_{\rm reg}\loss_{\rm reg}(\Theta),
  \qquad \lambda_m\geq 0.
  \label{eq:general-mixture-loss}
\end{equation}
Representative terms include the following.

\paragraph{Sampled Frobenius reconstruction.}
For very large tensors, evaluate only a mini-batch $\Omega$ of entries:
\begin{equation}
  \loss_{\Omega}(\Theta)
  = \frac{1}{2|\Omega|}\sum_{\vect{i}\in\Omega}
    \left(\widehat W_{\vect{i}}(\Theta)-W_{{\rm tar},\vect{i}}\right)^2.
  \label{eq:sampled-frob}
\end{equation}

\paragraph{Entrywise robust reconstruction.}
An $L_1$ or Huber term can reduce sensitivity to outlier weights:
\begin{equation}
  \loss_{1}(\Theta)
  = \sum_{\vect{i}}\left|\widehat W_{\vect{i}}(\Theta)-W_{{\rm tar},\vect{i}}\right|.
  \label{eq:l1-loss}
\end{equation}
AD frameworks use a valid subgradient or a smooth surrogate at zero.

\paragraph{Distillation and KL divergence.}
If $\vect{p}_{\rm T}$ is a teacher probability vector and $\vect{p}_\Theta$ is
the compressed model probability vector, then
\begin{equation}
  \loss_{\rm KL}(\Theta)
  = \sum_c p_{{\rm T},c}\log\frac{p_{{\rm T},c}+\delta}{p_{\Theta,c}+\delta},
  \qquad \delta>0.
  \label{eq:kl-loss}
\end{equation}
This term can be combined with Eq.~\eqref{eq:ce-loss} to preserve teacher
behaviour beyond top-1 accuracy \citep{Hinton2015}.

\paragraph{Regularisation, sparsity, and quantisation.}
Useful regularisers include weight decay, the soft isometry penalty in
Eq.~\eqref{eq:isometry-penalty}, gate sparsity, activation scale control, and
quantisation-aware penalties.  A practical objective is therefore
\begin{equation}
\begin{split}
  \loss(\Theta)
  ={}& \lambda_{\rm F}\loss_{\rm Frob}
     + \lambda_{\Omega}\loss_{\Omega}
     + \lambda_{\rm CE}\loss_{\rm CE}
     + \lambda_{\rm KL}\loss_{\rm KL} \\
     &+ \lambda_{\rm iso}\loss_{\rm iso}
     + \lambda_{\rm sp}\loss_{\rm sp}
     + \lambda_{\rm q}\loss_{\rm q}.
\end{split}
\label{eq:full-loss}
\end{equation}
The training algorithms below do not hard-code any one of these losses.  They
execute the chosen generator and downstream model, evaluate the scalar
$\loss(\Theta)$, and use reverse-mode AD to compute $\nabla_\Theta\loss$.

\section{Automatic Differentiation and Differentiable Programming}
\label{sec:ad}

\subsection{Efficient Gradient Computation}
\label{sec:grad}

Automatic differentiation (AD) represents a program as elementary differentiable
operations and applies the chain rule to the executed computation
\citep{Baydin2018,Domke2010,Halim2024,Holm2024}.  It produces
derivatives exact up to floating-point round-off, unlike finite differences,
which are approximate and step-size dependent \citep{Baydin2018}.  For a
supervised objective
\begin{equation}
  \loss(\theta)
  = \frac{1}{N}\sum_{i=1}^{N}
    \ell\!\left(f_\theta(\vect{x}_i),\vect{y}_i\right),
  \label{eq:loss-def}
\end{equation}
training requires $\nabla_\theta\loss$ even when $\theta$ contains millions or
billions of parameters.  Reverse-mode AD computes this gradient at a cost that
is typically a small constant multiple of the forward execution, assuming that
intermediate values are stored or recomputed by checkpointing \citep{Baydin2018}.

\subsection{Computational Graphs and Elementary Operations}
\label{sec:dag}

Let a program produce intermediate variables $v_j=f_j(v_{{\rm pa}(j)})$ on a
computational directed acyclic graph (DAG).  For a scalar loss $\loss$, define
the adjoint of any variable by
\begin{equation}
  \bar v_j := \frac{\partial \loss}{\partial v_j}.
  \label{eq:adjoint-def}
\end{equation}
Reverse-mode AD applies the local vector-Jacobian product
\begin{equation}
  \bar v_i
  \leftarrow
  \bar v_i +
  \left(\frac{\partial f_j}{\partial v_i}\right)^{\!*}\bar v_j,
  \qquad i\in {\rm pa}(j),
  \label{eq:local-vjp}
\end{equation}
for all nodes in reverse topological order.  This is the precise sense in which
backpropagation is the chain rule applied to the executed program.

For a fully connected layer,
\begin{equation}
  \vect{h}^{(\ell)}
  = \sigma\!\left(\mat{W}^{(\ell)}\vect{h}^{(\ell-1)}+\vect{b}^{(\ell)}\right),
  \label{eq:fc-layer}
\end{equation}
the graph contains matrix-vector multiplication, addition, and a pointwise
nonlinearity.  Tensor-network generators are handled in exactly the same way:
contractions, reshapes, permutations, activations, and losses are all graph
nodes with local backward rules.

\subsection{Forward-Mode and Reverse-Mode AD}
\label{sec:modes}

AD has two complementary modes \citep{Baydin2018}.
\begin{itemize}[leftmargin=*]
  \item \textbf{Forward mode} propagates tangents.  Given a direction
        $\dot\theta$, it computes
        $\dot\loss=\nabla_\theta\loss\cdot\dot\theta$.
  \item \textbf{Reverse mode} propagates adjoints.  Starting with
        $\bar\loss=1$, it computes all parameter adjoints by repeated use of
        Eq.~\eqref{eq:local-vjp}.
\end{itemize}
Forward mode is efficient when the number of inputs is small; reverse mode is
efficient when the number of scalar outputs is small.  Since training normally
uses one scalar loss and many parameters, reverse mode is the natural choice.
Modern systems such as TensorFlow, PyTorch, and JAX provide reverse-mode AD,
checkpointing, compilation, batching, mixed precision, and custom gradients
\citep{Abadi2016,Paszke2019,Bradbury2018}.

\subsection{Advantages over Numerical and Symbolic Differentiation}
\label{sec:adv}

Compared with finite differences and symbolic differentiation, AD is attractive
because it
\begin{itemize}[leftmargin=*]
  \item avoids finite-difference truncation and cancellation errors;
  \item avoids symbolic expression blow-up by differentiating the executed
        program; and
  \item composes cleanly through arbitrary differentiable modules, including
        tensor contractions and downstream neural-network losses.
\end{itemize}
Recent analyses in scientific machine learning also emphasise that replacing AD
by finite differences can alter optimisation dynamics, not only derivative
accuracy \citep{Chen2025}.

\subsection{Differentiable Programming}
\label{sec:dp}

Differentiable programming treats whole programs as trainable maps.  Tensor
network contractions, iterative solvers, spectral routines, and neural-network
layers can be combined in one differentiable objective, provided that the
implemented primitives have valid local backward rules \citep{Liao2019,LiaoLiu2021}.
For ADNTNs this means that the same code can support Frobenius reconstruction,
end-to-end cross-entropy, distillation, quantisation-aware fine-tuning, and
hardware-aware penalties without deriving a new optimisation algorithm for each
loss.

\section{Automatic-Differentiation Training Algorithms for Fundamental Nonlinear Tensor Networks}
\label{sec:training}

Classical TT-SVD, HOSVD \citep{DeLathauwer2000}, alternating least squares, and
DMRG-style updates rely on multilinearity: after all other cores are fixed, the
model is linear in the selected core.  Once an activation is inserted between contractions, this local
linear subproblem disappears.  In general,
\begin{equation}
  \sigma(\mat{A}\mat{B})\neq \sigma(\mat{A})\sigma(\mat{B}),
  \label{eq:nonlinear-nondistributive}
\end{equation}
where $\sigma$ is applied elementwise.  A nonlinear tensor network should
therefore be optimised as a differentiable program: define a forward contraction
schedule, evaluate a scalar loss, and apply reverse-mode AD to that executed
program.

This does not mean that AD makes tensor contraction free.  The memory and time
costs are controlled by the contraction order, the largest materialised
intermediate, batching strategy, checkpointing, and whether the full generated
weight tensor is cached or sampled.

\subsection{Core Tensor Estimation via AD and Adam/SGD Optimisation}
\label{subsec:core_estimation}

Let $\Theta$ denote all trainable cores and latent tensors.  For any core
$\ten{C}\in\Theta$, the AD gradient is a tensor of the same shape,
\begin{equation}
  \ten{G}_{C,t}
  := \nabla_{\ten{C}}\loss(\Theta_t)
  = \frac{\partial \loss(\Theta_t)}{\partial \ten{C}}.
  \label{eq:core-gradient-general}
\end{equation}
For a local operation $\ten{Y}=F(\ten{X},\ten{C})$, the two reverse rules are
\begin{equation}
  \overline{\ten{X}} =
  \left(\frac{\partial F}{\partial \ten{X}}\right)^{\!*}\overline{\ten{Y}},
  \qquad
  \overline{\ten{C}} =
  \left(\frac{\partial F}{\partial \ten{C}}\right)^{\!*}\overline{\ten{Y}}.
  \label{eq:tensor-vjp}
\end{equation}
For a bilinear contraction, Eq.~\eqref{eq:tensor-vjp} becomes a contraction of
the forward activation with the output adjoint over all indices except the core
indices.  It is therefore better to write the gradient as a contracted
``environment'' than as a simple outer product.

With stochastic gradient descent,
\begin{equation}
  \ten{C}_{t+1}=\ten{C}_t-\eta\ten{G}_{C,t}.
  \label{eq:sgd-update}
\end{equation}
For AdamW, using elementwise products and division, the update is
\begin{align}
  \ten{M}_t &= \beta_1\ten{M}_{t-1}+(1-\beta_1)\ten{G}_{C,t},
  \label{eq:adam-m}\\
  \ten{S}_t &= \beta_2\ten{S}_{t-1}+(1-\beta_2)(\ten{G}_{C,t}\odot\ten{G}_{C,t}),
  \label{eq:adam-s}\\
  \widehat{\ten{M}}_t &= \ten{M}_t/(1-\beta_1^t),\qquad
  \widehat{\ten{S}}_t = \ten{S}_t/(1-\beta_2^t),
  \label{eq:adam-bias}\\
  \ten{C}_{t+1}
  &= (1-\eta\lambda)\ten{C}_t
     -\eta\,\frac{\widehat{\ten{M}}_t}{\sqrt{\widehat{\ten{S}}_t}+\varepsilon_{\rm opt}}.
  \label{eq:adamw-update}
\end{align}
Here $\lambda$ is decoupled weight decay \citep{KingmaBa2015,LoshchilovHutter2019}.
All cores are updated simultaneously after the backward pass.

\subsection{Tree Tensor Network (TTN) Decoder}

\textbf{Layerwise forward model.}
Let $\ten{W}^{(0)}\in\R^{d\times\cdots\times d}$ be a latent tensor of order
$N_0=Q_{\rm bot}$.  A binary TTN doubles the number of modes at each layer,
$N_\ell=2N_{\ell-1}$.  For layer $\ell$ and branch $k$, let
\begin{equation}
  \ten{V}^{(\ell,k)}\in\R^{d\times d\times d},
  \qquad V^{(\ell,k)}_{p,a,b},
  \label{eq:ttn-core-shape}
\end{equation}
where $p$ is the parent index and $(a,b)$ are child indices.  The linear
pre-activation is
\begin{equation}
  Z^{(\ell)}_{c_1\ldots c_{N_\ell}}
  = \sum_{\vect{p}\in[d]^{N_{\ell-1}}}
      W^{(\ell-1)}_{p_1\ldots p_{N_{\ell-1}}}
      \prod_{k=1}^{N_{\ell-1}}
      V^{(\ell,k)}_{p_k,c_{2k-1},c_{2k}}.
  \label{eq:ttn-forward}
\end{equation}
The layer output is
\begin{equation}
  \ten{W}^{(\ell)}=\phi_\ell(\ten{Z}^{(\ell)}),
  \qquad
  \phi_\ell=
  \begin{cases}
    \sigma, & \ell<L,\\
    \id, & \ell=L.
  \end{cases}
  \label{eq:ttn-activation}
\end{equation}
The generated tensor is $\widehat{\ten{W}}_\Theta=\ten{W}^{(L)}$.

\textbf{Adjoint equations.}
Given $\overline{\ten{W}}^{(\ell)}=\partial\loss/\partial\ten{W}^{(\ell)}$, the
pre-activation adjoint is
\begin{equation}
  \overline{\ten{Z}}^{(\ell)}
  = \overline{\ten{W}}^{(\ell)}\odot \phi_\ell'\!\left(\ten{Z}^{(\ell)}\right).
  \label{eq:ttn-preactivation-adjoint}
\end{equation}
The adjoint of the previous layer is
\begin{equation}
  \overline{W}^{(\ell-1)}_{p_1\ldots p_{N_{\ell-1}}}
  = \sum_{\vect{c}\in[d]^{N_\ell}}
      \overline{Z}^{(\ell)}_{c_1\ldots c_{N_\ell}}
      \prod_{k=1}^{N_{\ell-1}}
      V^{(\ell,k)}_{p_k,c_{2k-1},c_{2k}}.
  \label{eq:ttn-parent-adjoint}
\end{equation}
The exact gradient of branch core $\ten{V}^{(\ell,k)}$ is
\begin{equation}
\begin{split}
  \overline{V}^{(\ell,k)}_{p,a,b}
  ={}& \sum_{\substack{\vect{p}\in[d]^{N_{\ell-1}},\ \vect{c}\in[d]^{N_\ell}\\
             p_k=p,\ c_{2k-1}=a,\ c_{2k}=b}}
      W^{(\ell-1)}_{p_1\ldots p_{N_{\ell-1}}}
      \overline{Z}^{(\ell)}_{c_1\ldots c_{N_\ell}} \\
     &\times
      \prod_{\substack{h=1\\ h\neq k}}^{N_{\ell-1}}
      V^{(\ell,h)}_{p_h,c_{2h-1},c_{2h}}.
\end{split}
\label{eq:ttn-core-gradient}
\end{equation}
Equation~\eqref{eq:ttn-core-gradient} is the manual environment expression that
AD constructs automatically from the contraction trace.

\begin{algorithm}[H]
\caption{Reverse-mode AD training for a nonlinear TTN decoder.  The algorithm generates a tensor by repeated expansion, evaluates a scalar reconstruction or task loss, and lets AD compute the adjoints in Eqs.~\eqref{eq:ttn-preactivation-adjoint}--\eqref{eq:ttn-core-gradient}.}
\begin{algorithmic}[1]
\Require Target data or task loss; depth $L$; latent order $N_0$; activation $\sigma$; optimiser.
\State Initialise $\ten{W}^{(0)}$ and TTN cores $\{\ten{V}^{(\ell,k)}\}$.
\State $\Theta\gets\{\ten{W}^{(0)}\}\cup\{\ten{V}^{(\ell,k)}\}$.
\For{training step $t=1,2,\ldots$}
  \State Clear gradients for all $\Theta$.
  \For{$\ell=1$ to $L$}
    \State Compute $\ten{Z}^{(\ell)}$ by Eq.~\eqref{eq:ttn-forward}.
    \State Compute $\ten{W}^{(\ell)}$ by Eq.~\eqref{eq:ttn-activation}.
  \EndFor
  \State Insert $\widehat{\ten{W}}_\Theta=\ten{W}^{(L)}$ into the layer/model and evaluate $\loss(\Theta)$.
  \State Call reverse-mode AD; this applies Eqs.~\eqref{eq:ttn-preactivation-adjoint}--\eqref{eq:ttn-core-gradient} implicitly.
  \State Update all cores using SGD or AdamW.
\EndFor
\Ensure Optimised TTN parameters $\Theta$.
\end{algorithmic}
\end{algorithm}

\subsection{Augmented TTN (aTTN): The Simplified MERA Decoder}

The aTTN uses TTN expansions in the hidden layers and applies lateral
disentanglers only at the final physical layer.  This is a compromise between a
strict TTN and a full MERA: it improves boundary mixing while keeping most of
the tree loop-free.

For hidden layers $1\leq\ell<L$, the forward equations are
Eqs.~\eqref{eq:ttn-forward}--\eqref{eq:ttn-activation}.  At the final layer,
first compute the ordinary TTN expansion
\begin{equation}
  \widetilde W^{(L)}_{c_1\ldots c_M}
  = \sum_{\vect{p}\in[d]^{N_{L-1}}}
      W^{(L-1)}_{p_1\ldots p_{N_{L-1}}}
      \prod_{k=1}^{N_{L-1}}
      V^{(L,k)}_{p_k,c_{2k-1},c_{2k}},
  \qquad M=N_L=2N_{L-1}.
  \label{eq:attn-final-expansion}
\end{equation}
Then apply open-boundary disentanglers to neighbouring pairs
$(c_{2k},c_{2k+1})$:
\begin{equation}
  Z^{(L)}_{e_1\ldots e_M}
  = \sum_{\vect{c}\in[d]^M}
      B^{(L)}_{\vect{e},\vect{c}}
      \widetilde W^{(L)}_{c_1\ldots c_M},
  \label{eq:attn-final-mixing}
\end{equation}
where
\begin{equation}
  B^{(L)}_{\vect{e},\vect{c}}
  = \delta_{e_1c_1}\delta_{e_Mc_M}
    \prod_{k=1}^{N_{L-1}-1}
    U^{(L,k)}_{e_{2k},e_{2k+1},c_{2k},c_{2k+1}}.
  \label{eq:attn-boundary-operator}
\end{equation}
The generated tensor is $\widehat{\ten{W}}_\Theta=\ten{Z}^{(L)}$.

The adjoint of the unmixed final tensor is
\begin{equation}
  \overline{\widetilde W}^{(L)}_{\vect{c}}
  = \sum_{\vect{e}\in[d]^M} B^{(L)}_{\vect{e},\vect{c}}
    \overline Z^{(L)}_{\vect{e}}.
  \label{eq:attn-unmixed-adjoint}
\end{equation}
The disentangler gradient is
\begin{equation}
\begin{split}
  \overline U^{(L,k)}_{\alpha,\beta,a,b}
  ={}& \sum_{\substack{\vect{e},\vect{c}\in[d]^M\\
      e_{2k}=\alpha,\ e_{2k+1}=\beta,\ c_{2k}=a,\ c_{2k+1}=b}}
      \overline Z^{(L)}_{\vect{e}}\,
      \widetilde W^{(L)}_{\vect{c}}\,
      \delta_{e_1c_1}\delta_{e_Mc_M} \\
    &\times
      \prod_{\substack{h=1\\h\neq k}}^{N_{L-1}-1}
      U^{(L,h)}_{e_{2h},e_{2h+1},c_{2h},c_{2h+1}}.
\end{split}
\label{eq:attn-u-gradient}
\end{equation}
Finally, $\overline{\widetilde{\ten{W}}}^{(L)}$ is propagated through the final
TTN expansion using the TTN adjoint formulae in
Eqs.~\eqref{eq:ttn-parent-adjoint}--\eqref{eq:ttn-core-gradient}.

\begin{algorithm}[H]
\caption{Reverse-mode AD training for an augmented TTN decoder.  Hidden layers follow the TTN expansion, while the final layer applies boundary disentanglers before the scalar loss is evaluated and differentiated.}
\begin{algorithmic}[1]
\Require Target data or task loss; depth $L$; activation $\sigma$; optimiser.
\State Initialise $\ten{W}^{(0)}$, TTN cores $\{\ten{V}^{(\ell,k)}\}$, and boundary disentanglers $\{\ten{U}^{(L,k)}\}$.
\State $\Theta\gets\{\ten{W}^{(0)},\ten{V}^{(\ell,k)},\ten{U}^{(L,k)}\}$.
\For{training step $t=1,2,\ldots$}
  \State Clear gradients.
  \For{$\ell=1$ to $L-1$}
    \State Apply TTN expansion and activation.
  \EndFor
  \State Compute final expansion by Eq.~\eqref{eq:attn-final-expansion}.
  \State Apply boundary disentanglers by Eq.~\eqref{eq:attn-final-mixing}.
  \State Evaluate $\loss(\Theta)$ and call reverse-mode AD.
  \State Update all cores with SGD or AdamW.
\EndFor
\Ensure Optimised aTTN parameters $\Theta$.
\end{algorithmic}
\end{algorithm}

\subsection{Nonlinear MERA Hierarchical Decoder}

A MERA-like decoder interleaves expansion tensors and disentanglers at every
scale.  For each layer $\ell$, first expand as in a TTN,
\begin{equation}
  \widetilde W^{(\ell)}_{c_1\ldots c_{N_\ell}}
  = \sum_{\vect{p}\in[d]^{N_{\ell-1}}}
      W^{(\ell-1)}_{p_1\ldots p_{N_{\ell-1}}}
      \prod_{k=1}^{N_{\ell-1}}
      V^{(\ell,k)}_{p_k,c_{2k-1},c_{2k}}.
  \label{eq:mera-expansion}
\end{equation}
Then mix adjacent boundary pairs using
\begin{equation}
  Z^{(\ell)}_{\vect{e}}
  = \sum_{\vect{c}\in[d]^{N_\ell}}
      B^{(\ell)}_{\vect{e},\vect{c}}
      \widetilde W^{(\ell)}_{\vect{c}},
  \label{eq:mera-mixing}
\end{equation}
with
\begin{equation}
  B^{(\ell)}_{\vect{e},\vect{c}}
  = \delta_{e_1c_1}\delta_{e_{N_\ell}c_{N_\ell}}
    \prod_{k=1}^{N_{\ell-1}-1}
    U^{(\ell,k)}_{e_{2k},e_{2k+1},c_{2k},c_{2k+1}}.
  \label{eq:mera-boundary-operator}
\end{equation}
The layer output is again
\begin{equation}
  \ten{W}^{(\ell)}=\phi_\ell(\ten{Z}^{(\ell)}),
  \qquad \phi_\ell=\sigma\ (\ell<L),\quad \phi_L=\id.
  \label{eq:mera-activation}
\end{equation}

In reverse mode,
\begin{align}
  \overline{\ten{Z}}^{(\ell)}
    &= \overline{\ten{W}}^{(\ell)}\odot\phi_\ell'(\ten{Z}^{(\ell)}),
    \label{eq:mera-z-adjoint}\\
  \overline{\widetilde{\ten{W}}}^{(\ell)}
    &= (D_{\ten{U}^{(\ell)}})^{*}\overline{\ten{Z}}^{(\ell)},
    \label{eq:mera-d-adjoint}\\
  \overline{\ten{W}}^{(\ell-1)}
    &= (E_{\ten{V}^{(\ell)}})^{*}\overline{\widetilde{\ten{W}}}^{(\ell)},
    \label{eq:mera-e-adjoint}
\end{align}
where $D$ denotes the mixing operator in Eq.~\eqref{eq:mera-mixing} and $E$
denotes the expansion operator in Eq.~\eqref{eq:mera-expansion}.  The gradients
of $\ten{U}^{(\ell,k)}$ and $\ten{V}^{(\ell,k)}$ are obtained by the same
contracted-environment forms as Eqs.~\eqref{eq:attn-u-gradient} and
\eqref{eq:ttn-core-gradient}, with $L$ replaced by $\ell$.

\begin{algorithm}[H]
\caption{Reverse-mode AD training for a nonlinear MERA decoder.  Each generator layer expands the state, mixes neighbouring branches with disentanglers, applies the layer activation, and then relies on reverse-mode AD to update all cores jointly.}
\begin{algorithmic}[1]
\Require Target data or task loss; depth $L$; activation $\sigma$; optimiser.
\State Initialise $\ten{W}^{(0)}$, expansion cores $\{\ten{V}^{(\ell,k)}\}$, and disentanglers $\{\ten{U}^{(\ell,k)}\}$.
\State $\Theta\gets\{\ten{W}^{(0)},\ten{V}^{(\ell,k)},\ten{U}^{(\ell,k)}\}$.
\For{training step $t=1,2,\ldots$}
  \State Clear gradients.
  \For{$\ell=1$ to $L$}
    \State Expand by Eq.~\eqref{eq:mera-expansion}.
    \State Mix by Eq.~\eqref{eq:mera-mixing}.
    \State Apply Eq.~\eqref{eq:mera-activation}.
  \EndFor
  \State Evaluate the scalar objective and call reverse-mode AD.
  \State Optionally add soft isometry, scale, sparsity, or quantisation penalties.
  \State Update all cores with SGD or AdamW.
\EndFor
\Ensure Optimised MERA parameters $\Theta$.
\end{algorithmic}
\end{algorithm}

\subsection{Advantages of AD vs. Classical Algorithms (SVD, DMRG, HOSVD)}
\label{sec:ad_vs_classical}

\textbf{Nonlinear generators.}
SVD, HOSVD, ALS, and standard DMRG updates exploit multilinearity.  AD only
requires differentiable primitives, so it can train SWISH/PELU/SIREN/rational/SwiGLU
ADNTNs without deriving a new environment equation for each activation.

\textbf{Task-aware losses.}
Classical decompositions most naturally minimise Frobenius reconstruction error.
AD can optimise cross-entropy, KL distillation, calibration, robustness,
quantisation, or hardware penalties directly.

\textbf{Correct treatment of contraction complexity.}
AD does not make exact contraction of arbitrary loopy tensor networks easy.  It
computes exact gradients of the contraction program that is actually executed.
For TTNs this program is naturally tree-structured.  For aTTN and MERA, the
chosen open-boundary expansion-mixing schedule and bounded local dimensions keep
intermediates manageable in the studied architectures.  More general PEPS-like
loopy networks would still require approximate contraction, sampling, or careful
contraction-order optimisation \citep{Ran2020}.

\subsection{Training Pipeline}
\label{sec:pipeline}

A robust ADNTN training schedule is as follows.
\begin{enumerate}[leftmargin=*]
  \item \textbf{Select layers.}  Compress the largest linear, convolutional, or attention projection layers first.
  \item \textbf{Tensorise.}  Choose radix $d$, mode ordering, latent order $Q_{\rm bot}$, bond dimension $\chi$, and topology.
  \item \textbf{Initialise consistently.}  Use the global policy in Section~\ref{sec:init}: He/Kaiming scaling for hidden nonlinear cores, Xavier/Glorot scaling for final linear cores, identity-plus-noise for disentanglers, and optional TT-SVD/Tucker warm starts.
  \item \textbf{Block-wise fitting.}  Fit one block or group at a time by reconstruction, input-output matching, or task loss.
  \item \textbf{End-to-end fine-tuning.}  Insert generated weights into the model and optimise the composite loss in Eq.~\eqref{eq:full-loss}.
  \item \textbf{Adaptive growth.}  Increase $\chi$, $d$, or selected disentangler layers where validation loss or gradient saliency indicates underfitting.
  \item \textbf{Quantisation-aware training.}  Add fake quantisation or quantisation penalties after the floating-point model has stabilised.
  \item \textbf{Prune, cache, and export.}  Remove small gates, fold constants, cache generated weights when appropriate, and export fused kernels.
\end{enumerate}

\section{Practical Implementations of ADNTNs and Computer Simulation Results}
\label{sec:experiments}

This section summarises the implementation choices and preliminary simulations.
Each compressed layer is implemented as a differentiable module: the generator
produces $\widehat{\mat{W}}_\Theta$ or $\widehat{\ten{K}}_\Theta$ in the forward
pass, the generated weight is used by the dense or convolutional operation, and
reverse-mode AD propagates gradients through both the operation and the
generator.  The experiments are intended as proof-of-concept evidence for the
ADNTN approach rather than a final benchmark suite.

\subsection{Grouped Weight Generation}
\label{sec:grouping}

Very large layers can be difficult to generate with one deep decoder.  We
therefore allow row grouping.  Partition $D_{\rm out}$ into $G$ groups of size
$D_{\rm block}=D_{\rm out}/G$.  Group $g$ has its own latent tensor and cores,
\begin{equation}
  \widehat{\mat{W}}^{(g)}
  = \operatorname{Generator}_g\!\left(\ten{W}^{(0,g)},\Theta_g\right)
  \in\R^{D_{\rm block}\times D_{\rm in}}.
  \label{eq:group-generator}
\end{equation}
The full matrix is reconstructed by row concatenation,
\begin{equation}
  \widehat{\mat{W}}_\Theta
  = \concat_{\rm rows}\!\left(
      \widehat{\mat{W}}^{(1)},\ldots,\widehat{\mat{W}}^{(G)}\right).
  \label{eq:group-concat}
\end{equation}
Grouping reduces generator depth, allows parallel independent generation, and
lets different row blocks specialise to different low-dimensional manifolds.  It
also introduces a modelling trade-off: cross-group interactions are not directly
parameterised inside the generator and must be captured by the task loss and
surrounding layers.

\subsection{Training Modality: End-to-end optimisation of task-aware loss functions}
\label{sec:training-modality}

In the experiments reported here, the compressed networks are trained from
scratch using the global initialisation policy in Section~\ref{sec:init}.  In
particular, the latent tensor is initialised as
$\ten{W}^{(0)}\sim\mathcal{N}(0,0.05^2)$, hidden expansion cores followed by SWISH/PELU
use He-style scaling, final linear readout cores use Xavier-style scaling, and
lateral disentanglers start near the identity map with small noise.  Using the
same policy for nonlinear TTN, aTTN, and MERA makes topology comparisons more meaningful
and reduces variance-related AD failures.  The primary objective is
cross-entropy, optimised with AdamW.  This end-to-end mode lets the model search
directly within the low-dimensional ADNTN manifold; in practice, the manifold
acts as a structural regulariser because many degrees of freedom that mainly
support overfitting are never introduced.

\subsection{Experimental Results}
\label{sec:experimental-results}

\textbf{Experimental setup.}
Unless otherwise stated, the reported simulations use a single group ($G=1$),
latent bottleneck order $Q_{\rm bot}=5$, physical radix and bond dimension
$d=\chi=2$, learnable SWISH or PELU activations between generator layers, the
global random initialisation of Section~\ref{sec:init}, reverse-mode AD,
cross-entropy loss, and AdamW.  Dense baselines are standard AlexNet- and
VGG-16-style models on the CIFAR-10 dataset
\citep{Krizhevsky2012,SimonyanZisserman2015}.

\textbf{Brick-wall ADTN reference.}
To make the comparison with \citet{Qing2025} concrete, we evaluate each ADNTN
topology against a brick-wall ADTN baseline that reproduces their published
design: four-index cores $\ten{A}^{[k]}\in\R^{2\times2\times2\times2}$, a fixed
boundary vector, fixed ReLU activations, and the same per-layer parameter budget.
The brick-wall entries in Table~\ref{tab:results}---$364$ generator parameters at
$81.09\%$ on the two largest AlexNet convolutional layers, and $1{,}820$
parameters at $91.09\%$ on the five largest VGG-16 convolutional layers---coincide
with the corresponding numbers reported by \citet{Qing2025}, so the rows labelled
``Brick Wall'' can be read directly as the ADTN baseline.  The hierarchical
generators are then trained under the identical data, optimiser, and seed
protocol, isolating the effect of topology and learnable nonlinearity.

Because the parameter counts and accuracy changes below are the central evidence
of this paper, the protocol that produces them must be fully specified.  For
each reported model we therefore fix and report (i)~the per-layer radix $d$ and
bond dimension $\chi$, the depth $L$, the branching schedule
$N_0,\ldots,N_L$ (Remark~\ref{rem:order-matching}), the number of disentangler
layers, and the group count $G$, so that $\bar P$ can be reconstructed from the
architecture; (ii)~the optimiser hyperparameters (learning rate, weight decay
$\lambda$, $\beta_1,\beta_2$, schedule), batch size, number of epochs, and data
augmentation; and (iii)~the number of random seeds, with the absolute dense
baseline accuracy alongside each $\Delta$.  The change $\Delta$ is meaningful
only relative to that baseline, and the standard deviations in
Table~\ref{tab:results} are computed over the same seeds.  The configuration
files, exact core shapes, and training code are released with the paper so that
every entry in Table~\ref{tab:results} is reproducible; the present numbers
should be read as a proof of concept on a small benchmark rather than a final
benchmark suite.

\textbf{Experiment 1: Compression of the two largest convolutional layers in AlexNet.}
AlexNet carries less redundancy than VGG-16 in this setup, so accuracy decreases
slightly under extreme compression---and it is precisely this low-redundancy
regime that most sharply separates the hierarchical generators from the brick
wall.  aTTN architecture was the most resilient hierarchical variant, losing  $-0.84$
percentage points accuracy at $2{,}656\times$ compression, while TTN and MERA lose $-1.52\%$
and $-0.63\%$ after compression, respectively.

\begin{figure}[H]
    \centering
    \incfig[0.8\textwidth]{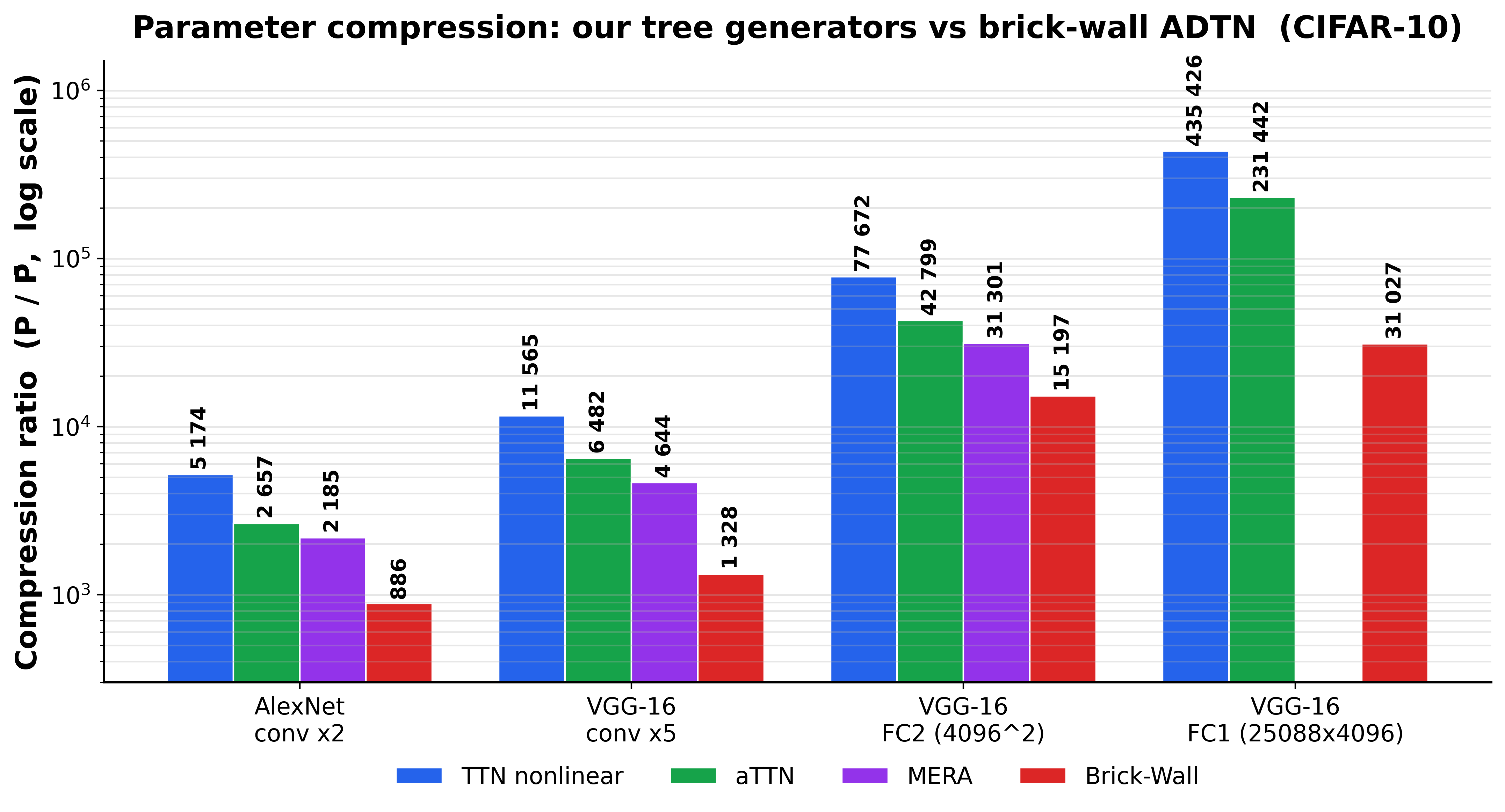}
    \caption{\textbf{Parameter-compression ratios $\rho_{\rm par}=P/\bar P$ for
    the selected compressed layers.}  The bars compare TTN, aTTN, MERA, and the
    brick-wall ADTN reference (assuming  three layers $M=3$ of brick-wall).  The plots compare  compression ratios archived for  the two largest AlexNet convolutional layers, the
    five largest VGG-16 convolutional layers, and the large VGG-16 fully connected
    layers FC2 and FC1.  The vertical axis is logarithmic; the ratios count
    trainable generator parameters only and should not be read as direct latency
    speedups. }
    \label{fig:compress}
\end{figure}

\begin{table}[H]
\centering
\caption{\textbf{Proof-of-concept CIFAR-10 results for ADNTN layer compression.}
Each row specifies the original model and layer group, the tensor-network
generator used to replace that group, the dense parameter count $P$, the
trainable generator parameter count $\bar P$, the parameter-compression ratio
$\rho_{\rm par}=P/\bar P$, and the validation accuracy with the change $\Delta$
(percentage points, p.p.) relative to the corresponding dense baseline.  The
dense baselines are $87.42\%$  for AlexNet (two convolutional layers) and  $90.19\%$ for 
VGG-16 (five convolutional layers and two fully connected
layers FC1/FC2).  The rows labelled ``Brick Wall'' reproduce the ADTN design of
\citet{Qing2025}  under the same protocol and serve as the reference baseline.
Reported deviations are standard deviations over repeated runs when available;
$\rho_{\rm par}$ is a compression-parameter ratio.}
\label{tab:results}
\resizebox{\textwidth}{!}{%
\begin{tabular}{l l l c r r r l}
\toprule
\textbf{Layer type} & \textbf{DNN model} & \textbf{TN arch.} & \textbf{Layers} &
$\boldsymbol{P}$ & $\boldsymbol{\bar P}$ & $\boldsymbol{\rho_{\rm par}}$ &
\textbf{Accuracy / $\Delta$} \\
\midrule
Conv   & AlexNet     & TTN        & 2 Conv & 1,572,864  & 304   & $5{,}173\times$  & $86.58\pm0.26\%$ / $-0.84$ p.p. \\
Conv   & AlexNet     & aTTN       & 2 Conv & 1,572,864  & 592   & $2{,}656\times$  & $85.90\pm0.20\%$ / $-1.52$ p.p. \\
Conv   & AlexNet     & MERA       & 2 Conv & 1,572,864  & 720   & $2{,}184\times$  & $86.79\pm0.34\%$ / $-0.63$ p.p. \\
Conv   & AlexNet     & Brick Wall & 2 Conv & 1,572,864  & 1780   & $886\times$  & $81.09\%$ / $-6.33$ p.p. \\
\midrule
Conv   & VGG-16      & TTN        & 5 Conv & 11,796,480 & 1,020 & $11{,}565\times$ & $91.38\pm0.11\%$ / $+1.19$ p.p. \\
Conv   & VGG-16      & aTTN       & 5 Conv & 11,796,480 & 1,820 & $6{,}481\times$  & $91.37\pm0.05\%$ / $+1.18$ p.p. \\
Conv   & VGG-16      & MERA       & 5 Conv & 11,796,480 & 2,540 & $4{,}644\times$  & $91.54\pm0.18\%$ / $+1.35$ p.p. \\
Conv   & VGG-16      & Brick Wall & 5 Conv & 11,796,480 & 9,216 & $1{,}280\times$  & $91.09\%$ / $+0.80$ p.p. \\
\midrule
Linear & VGG-16 FC2 & TTN        & 1 FC   & 16,777,216 & 216   & $77{,}672\times$ & $90.52\pm 0.19\%$ / $0.33$ p.p. \\
Linear & VGG-16 FC2 & aTTN       & 1 FC   & 16,777,216 & 392   & $42{,}799\times$ & $90.58\pm 0.11\%$ / $+0.39$ p.p. \\
Linear & VGG-16 FC2 & MERA       & 1 FC   & 16,777,216 & 536   & $31{,}301\times$ & $89.20\pm 0.24\%$ / $-0.99$ p.p. \\
\midrule
Linear & VGG-16 FC1 & TTN        & 1 FC   & 102,760,448 & 236   & $435{,}426\times$ & $90.59\pm 0.23\%$ / $+0.40$ p.p. \\
Linear & VGG-16 FC1 & aTTN       & 1 FC   & 102,760,448 & 444   & $231{,}442\times$ & $90.71\pm 0.18\%$ / $+0.52$ p.p. \\
\bottomrule
\end{tabular}%
}
\end{table}

\textbf{Experiment 2: Deep convolutional compression in VGG-16.}
Compressing the five largest VGG-16 convolutional layers yields strong results
across all three hierarchical generators.  At $11{,}565\times$ compression the
TTN model improves accuracy by $+1.19$ percentage points over the $90.19\%$ dense
baseline, and the MERA model reaches $+1.35$ points at $4{,}644\times$
compression.  Here VGG-16 is redundant enough that even the brick-wall ADTN
improves on the baseline ($+0.80$ points at $1{,}820$ parameters), but every
hierarchical generator still exceeds it---MERA by $+0.45$ points---at comparable
or smaller parameter budgets.  The low standard deviations ($\le 0.18\%$)
indicate that the optimisation is stable in this setting.

\begin{figure}[H]
    \centering
    \incfig[0.98\textwidth]{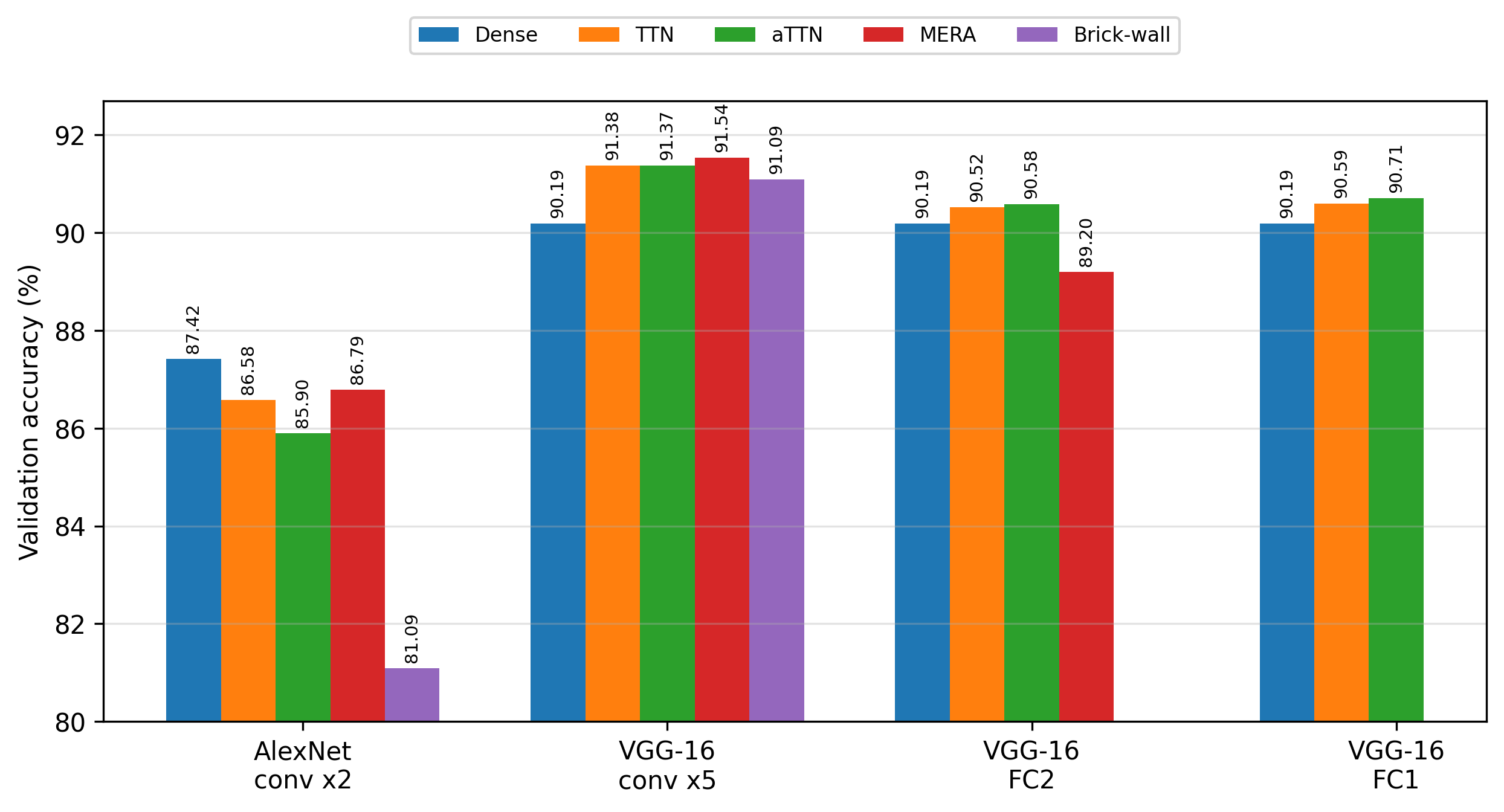}
    \caption{\textbf{Validation accuracy  of AlexNet and VGG-16 deep neural networks on CIFAR-10 after replacing selected
    fully connected or convolutional layers by ADNTN generators.}  The plot
    compares the dense baseline accuracy with  compressed DNNs via the TTN, aTTN, and MERA generators for the
    same layer groups as in Table~\ref{tab:results}.  On VGG-16 the compressed
    manifold often improves generalisation, whereas the more compact AlexNet
    baseline leaves less redundancy for extreme compression and shows a slight
    accuracy reduction.}
    \label{fig:accuracy}
\end{figure}

\textbf{Experiment 3: Compression of fully connected layer FC2 in VGG-16.}
The VGG-16 FC2 layer contains $4096\times4096=16{,}777{,}216$ parameters.  The TTN
version uses only $216$ trainable generator parameters, corresponding to
$77{,}672\times$ compression, with a  slight accuracy improvement of $+0.33$
percentage points relative to the $90.19\%$ baseline.  The aTTN version uses
$392$ parameters and improves accuracy by $+0.39$ points, indicating that the
boundary disentanglers add useful capacity at modest cost and that the ADNTN
manifold can act as a useful implicit regulariser even at $4\times10^{4}$-fold
compression, while MERA provide reduced accuracy by almost one percent.

\textbf{Experiment 4: Compression of the large fully connected layer FC1 in VGG-16.}
The VGG-16 FC1 layer contains $25088\times4096=1.0276\times10^{8}$ parameters---an
order of magnitude larger than the linear layers compressed by \citet{Qing2025},
and the most extreme regime tested here.  The TTN version uses only $236$
trainable generator parameters, corresponding to $435{,}426\times$ compression,
yet still improves accuracy by $+0.40$ percentage points over the $90.19\%$
baseline.  The aTTN version uses $444$ parameters ($231{,}442\times$ compression)
and improves accuracy by $+0.52$ points.  MERA was not tested extensively on this
huge layer because of its higher contraction cost and its slightly lower accuracy than
aTTN in the FC setting.  That two hundred-odd parameters can replace a hundred
million while \emph{improving} generalisation is the strongest single piece of
evidence that many trained DNN layers occupy a highly structured, low-dimensional
region of weight space that the ADNTN generator can reach for specific appliacations.
%

\subsection{Optimising DNN Inference Latency:\\ SRAM Implementation and Caching Strategies for Nonlinear Tensor Networks}
\label{sec:latency}

Extreme parameter compression does not automatically imply latency reduction.
If a generated dense weight is reconstructed on every inference call, the
generator can dominate runtime.  A deployment-oriented ADNTN implementation
therefore has two modes.
\begin{itemize}[leftmargin=*]
  \item \textbf{Cached-weight mode:} generate $\widehat{\mat{W}}_\Theta$ once,
        cache it, and run the original dense/convolutional kernel.  This is
        simplest and usually fastest when memory allows.
  \item \textbf{On-the-fly mode:} evaluate the tensor generator inside a fused
        kernel.  This is attractive when weights are too large to store or when
        inputs allow activation reuse, but it requires careful compilation.
\end{itemize}

\textbf{SRAM/cache residency.}
The core tensors in Table~\ref{tab:results} occupy from hundreds of bytes to a
few tens of kilobytes depending on precision.  Such parameter sets can often
reside in L1/L2 cache, GPU shared memory, or microcontroller SRAM/TCM.  In a
custom CUDA kernel, cores can be loaded into \texttt{\_\_shared\_\_} memory at
kernel start; on edge devices, cores can be allocated in tightly coupled memory
where available.  This reduces memory traffic, but the actual latency gain must
be measured on the target hardware.

\textbf{Activation reuse.}
For structured inputs with repeated local patterns, the first one or two tree
levels may repeatedly receive identical sub-vectors.  A small look-up table can
cache post-activation values of the form
$\sigma(\operatorname{Core}\times\operatorname{InputSubvector})$.  This is most
useful for sparse, padded, or repeated inputs; it should be disabled when cache
lookups exceed saved arithmetic.

\textbf{Graph compilation and fusion.}
Dozens of small contractions can be slower than one dense kernel if executed as
separate framework operations.  JAX \texttt{jit}, PyTorch \texttt{compile}, XLA,
and Triton-style custom kernels can fuse contractions, activations, and reshapes
so that intermediates remain in registers or shared memory \citep{Bradbury2018,Paszke2019}.
The implementation target is a small number of fused kernels per compressed
layer, not a Python loop over individual cores.

\textbf{Branch-level parallelism.}
In TTN and aTTN hidden layers, all branches at the same depth are independent.
They should be mapped to parallel threads, warps, or vector lanes.  MERA adds
lateral mixing, but contractions within a scale still expose useful parallelism.
In preliminary cached-weight implementations, this approach produced inference latencies in the range of $12$--$18$ ms/image; these numbers should be treated as hardware- and implementation-dependent rather than universal.

\begin{figure}[H]
    \centering
    \incfig[0.6\textwidth]{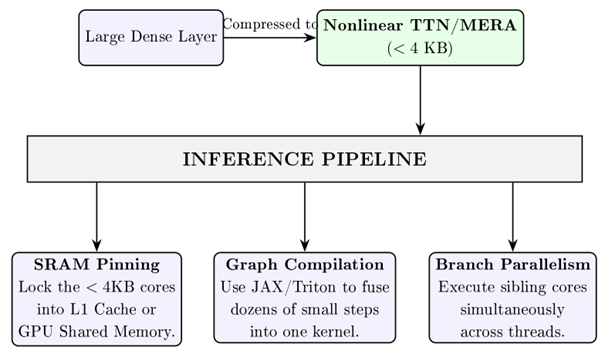}
    \caption{\textbf{Deployment pipeline for turning extreme parameter
    compression into practical inference speed.}  A large dense layer is replaced
    by a compact nonlinear TTN/MERA generator whose cores can be pinned in cache
    or SRAM.  Low latency then depends on implementation choices: cached weights
    when memory allows, fused on-the-fly generation when memory is the
    bottleneck, and branch-level parallelism to avoid many small sequential
    kernel launches.}
    \label{fig:latency-pipeline}
\end{figure}

The most realistic near-term deployment strategy is optimistic but pragmatic:
train ADNTNs end-to-end, cache generated weights when memory allows, and use
fused on-the-fly generation only for layers or devices where the memory savings
outweigh the additional arithmetic.

\section{Conclusions and Future Directions}
\label{sec:conclusions}

\subsection{Summary of Contributions}
\label{sec:summary}

This paper has developed ADNTNs as a principled, mathematically grounded framework
for extreme parameter compression of deep neural networks via multilayered hierarchical nonlinear tensor networks  
A central observation, simple yet consequential, drives the entire framework:
inserting learnable nonlinear activations between tensor contraction layers
destroys the multilinear structure on which SVD, HOSVD, TT-SVD, alternating
least squares (ALS), and DMRG-style local updates rely.  Automatic differentiation, by
contrast, operates on the execution trace rather than the spatial topology of the
network and requires only that every primitive in the forward program be
differentiable.  This makes automatic differentiation (AD) not merely a convenient implementation tool but
the \emph{necessary} algorithmic foundation for training nonlinear tensor-network
generators end-to-end against arbitrary task-aware objectives.

\paragraph{Theoretical.}
We made explicit the cut-rank restrictions that govern purely multilinear tensor
networks and explained why inserting nonlinear activations between contraction
layers  of core tensors changes the picture.  
The claim is deliberately cautious: tensor networks  with  small  core tensors  with ranks (bonds) $\chi=2$ are not always universal representations of all dense matrices.
Nevertheless, nonlinear generators can explore weight families that are much
richer than the corresponding multilinear tensor networks, and this matches the
practical hypothesis that many trained DNN layers occupy structured, highly
compressible regions of parameter space.

\paragraph{Architectural.}
We described nonlinear TTN, aTTN, and MERA decoders using the same notation for state
tensors, pre-activations, expansion cores, disentanglers, and generated weights.
TTN gives the simplest and most economical hierarchy; aTTN adds a targeted
boundary-mixing mechanism; MERA gives the strongest multi-scale mixing, at the
price of more permutations and larger contraction constants.  The revised
initialisation policy also makes these architectures easier to compare because
core variance, latent scale, and disentangler perturbations are handled in a
single consistent way.

\paragraph{Algorithmic.}
We derived forward-adjoint equations for the studied tensor network generators.  
These formulae
show that reverse-mode AD computes pre-activation adjoints, parent-state
adjoints, and contracted-environment gradients for every trainable core.  This
is valuable because it
lets nonlinearities, task-aware losses, batching, modern optimisers, and
hardware-aware schedules be handled inside one differentiable program.

\paragraph{Empirical.}
The proof-of-concept  experiments on AlexNet and VGG-16 layers are
encouraging and, in terms of huge compression ratio.  Per-layer compression
ratios range from roughly $2{,}000\times$ to $430{,}000\times$, and several
VGG-16 configurations match or marginally \emph{exceed} the dense baseline in
top-1 validation accuracy, suggesting that the ADNTN manifold acts as a
beneficial structural regulariser.  Under a matched protocol, all three
hierarchical generators outperform a faithfully reproduced brick-wall ADTN
baseline \citep{Qing2025} in respect compression ratio and similar or better accuracy.
 These
results should be read as proof of concept rather than a definitive benchmark:
they establish feasibility and motivate further investigation, but broader
validation on larger datasets, transformer architectures, and LLM projection
layers is essential before drawing stronger conclusions.

\subsection{Advantages of the AD Approach over Classical Methods}
\label{sec:ad_advantages_conclusion}

Relative to classical multilinear solvers such as SVD, HOSVD, ALS, and DMRG, the
automatic-differentiation formulation underlying ADNTNs offers four concrete
advantages.

\textbf{(i) Unrestricted nonlinear approximation.}
Because reverse-mode AD only traverses the computation graph by the chain rule,
it computes exact gradients through any differentiable activation $\sigma$.  This
lets the generator break the multilinear cut-rank bound of
Eq.~\eqref{eq:cut-rank-bound} and produce high-rank, structured weight tensors
$\widehat{\ten{W}}_\Theta$ from exponentially compressed cores at $\chi=2$, a
regime that classical multilinear factorisations and simple  tensor train (TT/MPS) 
or tensor ring (TR) cannot reach.

\textbf{(ii) Task-aware loss functions.}
SVD/HOSVD and DMRG minimise Euclidean (Frobenius) reconstruction error by
construction.  AD instead optimises the entire generator end-to-end against
cross-entropy, knowledge-distillation, calibration, or any other differentiable
objective.  By backpropagating task error directly into the cores, AD allocates
the scarce virtual-bond capacity to the parameter directions that matter most for
the downstream task rather than distributing it uniformly across all weight
entries.

\textbf{(iii) Stochastic sub-sampling.}
Approximating a massive target tensor does not require materialising it in full.
By querying the generator for a random batch of multi-indices, AD performs
stochastic gradient descent on the sampled reconstruction loss of
Eq.~\eqref{eq:sampled-frob}, circumventing the memory wall that limits dense
SVD/HOSVD and allowing the target order $Q$ to scale far beyond what explicit
factorisation permits.

\textbf{(iv) Tractable handling of lateral mixing.}
Exact contraction of loopy tensor networks is generally hard, and the
disentanglers of aTTN and MERA introduce exactly such lateral couplings.  AD
sidesteps the difficulty by differentiating the specific open-boundary
expansion--mixing schedule that is actually executed (Step~A: expansion cores;
Step~B: disentanglers), so the studied architectures yield exact gradients with
bounded intermediates without ever solving an undirected cyclic contraction.  As
noted in Section~\ref{sec:ad_vs_classical}, this is a statement about the executed
program, not a claim that AD makes arbitrary loopy contraction free.

\subsection{Limitations}
\label{sec:limitations}

While AD supplies the exact gradients required to train deep nonlinear
generators, it introduces distinct computational challenges that must be managed
in practice.

\begin{itemize}[leftmargin=*]
  \item \textbf{Optimisation remains non-convex.}  AD supplies accurate gradients
        of the executed program, not global optima.  Initialisation, activation
        scaling, optimiser choice, and learning-rate schedules materially affect
        the result.
  \item \textbf{Variance control matters.}  Deep generators with many small
        cores are sensitive to the initial scale of latent tensors, expansion
        cores, and disentanglers.  The global initialisation convention proposed
        here should therefore be treated as part of the model specification, not
        as an implementation afterthought.
  \item \textbf{Contraction and materialisation costs remain real.}  If a full
        generated weight is materialised at every training or inference step,
        memory and runtime can dominate.  Sampling, caching, checkpointing, and
        careful contraction ordering are important.
  \item \textbf{Latency is hardware-dependent.}  Parameter compression does not
        automatically imply wall-clock speedup.  Practical acceleration requires
        fused kernels, cache/SRAM residency, and hardware-specific measurement.
  \item \textbf{The evidence is preliminary.}  The reported simulations are
        promising but limited.  Larger datasets, transformer blocks, LLM
        projections, ablations, and reproducible implementations are important
        next steps.
  \item \textbf{Claims depend on a fully specified protocol.}  The headline
        compression ratios and accuracy changes are only interpretable together
        with the absolute dense baseline, the exact per-layer architecture that
        determines $\bar P$ (Remark~\ref{rem:order-matching}), and the number of
        seeds.  We treat the released configuration and code as part of the
        result, not as supplementary detail; single-point comparisons without an
        accuracy/compression Pareto sweep can over- or under-state the benefit.
\end{itemize}

\subsection{Future Directions}
\label{sec:future}

ADNTNs are best understood as trainable weight generators rather than fixed
post-hoc decompositions.  That viewpoint opens several constructive research
directions.
\begin{itemize}[leftmargin=*]
  \item \textbf{Adaptive topology search.}  Learn mode ordering, tree
        connectivity, bond dimensions, and disentangler placement jointly with
        the core tensors.
  \item \textbf{Faithful compression.}  Combine cross-entropy with knowledge
        distillation, feature matching, attention transfer, calibration losses,
        and safety-relevant behavioural tests so that the compressed model
        preserves more than top-1 accuracy.
  \item \textbf{Hardware-aware co-design.}  Optimise contraction order,
        precision, caching strategy, quantisation format, and kernel fusion with
        hardware-in-the-loop feedback.
  \item \textbf{Dynamic growth and pruning.}  Grow $\chi$, $d$, or selected
        disentangler layers only where validation loss or gradient saliency
        indicates underfitting, and prune gates or cores where sparsity penalties
        indicate redundancy.
  \item \textbf{Operator learning.}  Use encoder-decoder TTN/MERA topologies as
        compact nonlinear operators for images, signals, and scientific-computing
        surrogates.
  \item \textbf{Quantum and neuromorphic links.}  MERA-like causal structures
        connect naturally to quantum circuits and specialised neuromorphic
        accelerators; as hardware evolves, such generators may become even more
        attractive, particularly for models designed to run on near-term quantum
        devices.
  \item \textbf{Scaling to LLMs.}  Applying ADNTNs to the attention projection
        matrices and FFN layers of large language models is a natural and
        high-impact next step.  The joint QKV tensorisation introduced in
        Section~\ref{sec:Tensorization} provides a starting point, but
        systematic ablation studies on models such as LLaMA, GPT-2, and Mistral
        are needed to assess whether the compression ratios observed for
        AlexNet/VGG-16 carry over to the transformer setting.
  \item \textbf{Theoretical foundations.}  Establishing formal approximation
        guarantees for specific weight-tensor families---for example, proving
        that attention weight matrices of transformer models trained on
        structured tasks lie close to the ADNTN manifold for moderate $\chi$---
        would place the compression claims of this paper on a firmer theoretical
        footing.
\end{itemize}

The overall message is optimistic but rigorous.  The transition from classical
tensor algorithms (SVD, HOSVD, DMRG) to automatic differentiation is not a mere
software convenience---it is a \emph{mathematical and conceptual necessity} for
deep nonlinear generators.  ADNTNs do not remove the need for careful
mathematics, stable initialisation, and disciplined experimental protocols; the
classical tensor network algorithms (SVD, HOSVD, TT-SVD) remain valuable as
initialisation tools and analytical benchmarks.  What ADNTNs offer beyond
classical methods is a coherent way to combine tensor-network structure with
modern differentiable programming, arbitrary task-aware loss functions, and the
full ecosystem of GPU-optimised AD frameworks.  If future experiments on larger
architectures confirm the early large compression ratios and generalisation
trends reported here, ADNTNs---possibly combined with existing classical tensor networks algorithms could become a
credible path toward deep neural networks that are far smaller, easier to deploy, and
still expressive enough for demanding specific tasks.

\end{document}